%% file: ms.tex
\documentclass{article}

\usepackage[utf8]{inputenc}
\usepackage[english]{babel}
\usepackage[T1]{fontenc}

\usepackage{microtype}
\usepackage{graphicx}
\usepackage{subcaption}
\usepackage{booktabs}
\usepackage{multirow}
\usepackage{multicol}
\usepackage{enumitem}
\usepackage{placeins}
\usepackage{balance}

\usepackage[accepted]{icml2019}

\usepackage[breaklinks=true]{hyperref}

\input{my-definitions}

\graphicspath{{figures/}}

\icmltitlerunning{Are Graph Neural Networks Miscalibrated?}

\begin{document}

\twocolumn[
\icmltitle{Are Graph Neural Networks Miscalibrated?}




\begin{icmlauthorlist}
\icmlauthor{Leonardo Teixeira}{purdue}
\icmlauthor{Brian Jalaian}{arl}
\icmlauthor{Bruno Ribeiro}{purdue}
\end{icmlauthorlist}

\icmlaffiliation{purdue}{Department of Computer Science, Purdue University, West Lafayette, IN, USA}
\icmlaffiliation{arl}{U.S. Army Research Laboratory, Adelphi, MD, USA}

\icmlcorrespondingauthor{Leonardo Teixeira}{lteixeir@purdue.edu}

\icmlkeywords{Machine Learning, Relational Learning, Calibration, Uncertainty, Deep Learning}

\vskip 0.3in
]



\printAffiliationsAndNotice{}  


\begin{abstract}
Graph Neural Networks (GNNs) have proven to be successful in many classification tasks, outperforming previous state-of-the-art methods in terms of accuracy.
However, accuracy alone is not enough for high-stakes decision making.
Decision makers want to know the likelihood that a specific GNN prediction is correct.
For this purpose, obtaining calibrated models is essential.
In this work, we perform an empirical evaluation of the calibration of state-of-the-art GNNs on multiple datasets.
Our experiments show that GNNs can be calibrated in some datasets but also badly miscalibrated in others, and that state-of-the-art calibration methods are helpful but do not fix the problem.
\end{abstract}


\input{introduction}

\input{background}
\input{miscalibration}
\input{conclusion}


 \section*{Acknowledgements}
This work was sponsored in part by the ARO, under the U.S. Army Research Laboratory contract   number W911NF-09-2-0053, the Purdue Integrative Data Science Initiative and the Purdue  Research foundation, the DOD through SERC under contract number HQ0034-13-D-0004RT \#206, and the National Science Foundation under contract numbers IIS-1816499 and DMS-1812197.

\FloatBarrier
\bibliography{ms}
\bibliographystyle{icml2019}





\input{appendix}

\end{document}

%% file: my-definitions.tex
\usepackage[usenames,dvipsnames]{xcolor}
\usepackage{mathtools}  
\usepackage{amsfonts}
\usepackage{amssymb}
\usepackage{amsthm}
\usepackage{dsfont}
\usepackage{bm}
\usepackage{mathrsfs}

\usepackage{physics}
\usepackage{pbox}
\usepackage{xparse}
\usepackage{xspace}

\usepackage[nameinlink]{cleveref}

\newcommand{\Appendix}{Supplementary Material\xspace}

\newtheorem{definition}{Definition}

\theoremstyle{remark}

\NewDocumentCommand{\VertexSet}{}{V}
\NewDocumentCommand{\EdgeSet}{}{E}

\NewDocumentCommand{\Graph}{}{G}
\NewDocumentCommand{\Neighbors}{m}{\mathcal{N}(#1)}

\NewDocumentCommand{\NumClasses}{}{C}

\NewDocumentCommand{\ECEOdds}{}{\ensuremath{\text{ECE}_{\geq 50}}\xspace}

\NewDocumentCommand{\feature}{so}{\IfBooleanTF{#1}{x}{X}\IfValueT{#2}{_{#2}}}
\NewDocumentCommand{\target}{so}{\IfBooleanTF{#1}{y}{Y}\IfValueT{#2}{_{#2}}}

\NewDocumentCommand{\predicted}{so}{\hat{\IfBooleanTF{#1}{Y}{y}}\IfValueT{#2}{_{#2}}}
\NewDocumentCommand{\confidence}{o}{\hat{p}\IfValueT{#1}{_{#1}}}
\NewDocumentCommand{\correct}{o}{c\IfValueT{#1}{_{#1}}}

\NewDocumentCommand{\given}{}{\mid}

\NewDocumentCommand{\dataset}{o}{\mathcal{D}\IfValueT{#1}{^{(\text{#1})}}}

\NewDocumentCommand{\model}{o}{f\IfValueT{#1}{_{#1}}}
\NewDocumentCommand{\Ind}{r[]}{\ensuremath{\mathds{1}\left[#1\right]}}

\NewDocumentCommand{\Param}{ s o d() }{
    \def\myparam{
        \IfBooleanF{#1}{
            \IfNoValueT{#2}{\bm}
        }
        {\theta}
    }
    \myparam\IfValueT{#2}{_{#2}}
    \IfValueT{#3}{^{(#3)}}
}

\NewDocumentCommand{\Prob}{o d()}{\ensuremath{
    \mathds{P}
    \IfValueT{#1}{
        ^{\text{(#1)}}
    }
    \IfValueT{#2}{(
        {#2}
    )}
}}

\NewDocumentCommand{\Exp}{d() r[]}{\ensuremath{
    \mathds{E}
    \IfValueT{#1}{_{#1}}
    {\left[#2\right]}
}}

\NewDocumentCommand{\presoftmax}{od()}{
    \IfValueTF{#1}{\vh_{#1}}{\mH}
    \IfValueT{#2}{^{(#2)}}
}

\NewDocumentCommand{\postsoftmax}{od()}{
    \IfValueTF{#1}{\vp_{#1}}{\mP}
    \IfValueT{#2}{^{(#2)}}
}

\NewDocumentCommand{\Reals}{o}{
    \mathds{R}
    \IfValueT{#1}{
    ^{#1}
    }
}

\def\1{\bm{1}}

\def\vh{{\bm{h}}}

\def\vp{{\bm{p}}}

\def\mH{{\bm{H}}}

\def\mP{{\bm{P}}}

\DeclareMathAlphabet{\mathsfit}{\encodingdefault}{\sfdefault}{m}{sl}
\SetMathAlphabet{\mathsfit}{bold}{\encodingdefault}{\sfdefault}{bx}{n}

\DeclareMathOperator*{\argmax}{arg\,max}


%% file: introduction.tex
\section{Introduction}
\label{sec:introduction}

Modern graph neural networks (GNNs) are more accurate than previous state-of-the-art models 
and have proven useful in a host of supervised learning tasks over relational data~\cite{Battaglia2018} including visual scene understanding \citep{Raposo2017}, few-shot learning \citep{Satorras2018}, learning dynamics of physical systems \citep{Sanchez-Gonzalez2018}, learning multiagent communications \citep{Sukhbaatar2016}, predicting chemical properties of molecules \citep{Duvenaud2015,Gilmer2017}, to name a few.

However, raw accuracy measures are not enough in high-stakes decision-marking: stakeholders need to know {\em which predictions should they really trust and which ones are likely unreliable.}
Models with softmax outputs ---and trained with cross-entropy and likelihood losses--- are able to output a probability that the predicted label is indeed the correct answer. 
{\em But can we trust these softmax probabilities?}
At the core of this question lies the principle of calibration.
%
In a \emph{calibrated} model the softmax output of the predicted label actually matches the relative frequency that the prediction is correct, {\em i.e.}, if the softmax output of the predicted label gives 0.8, then 8 out of 10 times the label is correct.
Having a calibrated model is an essential requirement for any decision-making task.

Calibration (a.k.a.\ reliability) is a property of uncertainty both in the model parameters and in the model itself (mispecification), and as such, it is a challenge for both frequentist and Bayesian models alike~\cite{rubin1984bayesianly}.
Calibration is an important tool to assess the quality of the model predictions, from the point of view of reliably estimating its uncertainty.
It is also a metric that is orthogonal to model accuracy ---a classifier whose predictions are random (from the class priors) will be perfectly calibrated.

\paragraph{Are GNNs calibrated?}%
The literature on calibration is missing a thorough evaluation of the calibration of GNNs, which consider dependent inputs (relational data), in contrast to traditional objectives that consider independent and identically distributed (i.i.d.) data.
This work investigates the (mis)calibration of GNNs, and how techniques commonly used for calibration of i.i.d.\ data will perform in GNNs over non-i.i.d.\ (relational) data.
Our experiments show that, while state-of-the art calibration methods can be useful, for some harder tasks they do not solve the problem, from which we conclude that \emph{GNNs can be miscalibrated and existing calibration methods cannot fix it}.

\paragraph{Contributions.}%
Our main contributions are: 
(1) empirical evaluation of the calibration of GNNs on frequently used graph datasets; 
and (2) showing that simple and state-of-the-art calibration methods are not enough to calibrate GNNs.

\paragraph{Related Work.}%
Calibration has been extensively studied in the context of classical machine learning tasks, such as binary classification \citep{Zadrozny2001, Zadrozny2002, Niculescu-Mizil2005, Platt1999, Gao2017}, and in classical statistical tasks~\citep{rubin1984bayesianly,box1980sampling}.
Recently, \citet{Guo2017} shows how modern neural networks (in contrast to simpler architectures of decades ago), while very accurate, are also miscalibrated; and how a simple technique, called \emph{temperature scaling} is an effective method to calibrate image classifiers (convolutional neural networks).
In the context of deep learning models for regression tasks, \citet{Kuleshov2018} recently proposed a simple calibration method, based on isotonic regression.

Since then, temperature scaling remained the go-to calibration method for deep learning models, while other works have investigated improvements with better loss functions \citep{Kumar2018, Mozafari2018}.
While there are other works investigating improvement of uncertainty quantification in deep learning \citep{Gal2016,Lakshminarayanan2017,Card2019}, they neither target nor investigate calibration. 
In particular, to the best of our knowledge, the calibration of graph neural networks has not been investigated.

%% file: background.tex
\section{Background and definitions}
\label{sec:background}

\paragraph{Graph Neural Networks (GNNs).}%
Consider a graph \(\Graph=(\VertexSet, \EdgeSet)\) where \(\VertexSet\) is the set of \(n\) vertices (or nodes), \(\EdgeSet\) is the set of edges. 
Each node \(u \in \VertexSet\) has an associated vector of attributes \(\feature[u]^{(attr)} \in \Reals[d]\). 
Denote by \(\Neighbors{u}\) the set of neighbors of node \(u\).

GNNs~\citep{Kipf2017, Velickovic2018, Xu2018} (among others) are neural network models where, for each layer \(k\), a hidden representation for the node \(u\) is computed based on previous representation of neighboring nodes, as follows:
\begin{gather*}
    h_u^{(k)} = \phi^{(k)}\left(h_u^{(k-1)}, h_{\Neighbors{u}}^{(k)}\right), \\
    h_{\Neighbors{u}}^{(k)} = \rho^{(k)}(\{h_j^{(k-1)} : j \in \Neighbors{u} \}),
\end{gather*}
where usually \(h_u^{(0)} = \feature[u]^{(attr)}\), \(\phi^{(k)}\) is a non-linear transformation and \(\rho^{(k)}\) is a pooling operator, e.g. sum, mean or more powerful LSTM-type aggregators~\citep{Murphy2019a}. 
After \(K\) layers, the final embedding \(h_u^{(K)}\) is obtained. 
Then, a softmax is applied to produce probabilities, which are used to predict the node's class. 
The models are trained by minimizing the negative log-likelihood through gradient descent.
In an abuse of notation, we use \(\feature[u]\) to denote all variables needed to compute \(h_u^{(K)}\), i.e. features and edges within a ball of radius $K$ around $u$. 

\paragraph{Model Calibration.}%
Consider a model \(\model[\Param]\), parameterized by \(\Param\) trained for a classification task, where its input is denoted by \(\feature\) and the target class label by \(\target \in \{1, 2, \dots, \NumClasses\}\).
For a given input \(i\), denote by \(\predicted[i]\) the predicted label:
\begin{align*}
\predicted[i] = \predicted(\feature[i]) = \argmax_{k\in \{1, \dots, \NumClasses\}} \left[\model[\Param](\feature[i])\right]_{k},
\intertext{and \(\confidence[i]\) its predicted probability (confidence):}
\confidence[i] = \confidence(\feature[i]) = \max_{k\in \{1, \dots, \NumClasses\}} \left[\model[\Param](\feature[i])\right]_{k}. 
\end{align*}
\begin{definition}\label{def:calib}
The classifier $\model[\Param]$ is said to be calibrated iff \(\Prob(\target = \predicted(\feature) \given \confidence(\feature) = p ) = p, \;\; \forall p \in \left[\frac{1}{C}, 1 \right]\), where \(C\) is the number of classes.
\end{definition}
Note that for this definition of calibration, only the predicted class is taken into account. 
While this is the most common definition for deep learning, other definitions of calibration are possible, with different implications \citep{Vaicenavicius2019}.
Intuitively, \Cref{def:calib} means that the confidence of the predictions should match the frequency that they are correct.
For example: if, among the predictions made by the model, there are 100 predictions made with confidence of \(0.7\), we would expect \(70\) of them to be correct.

\paragraph{Evaluating Calibration.}%
We employ the two common tools used in the literature to evaluate calibration: \emph{reliability diagrams}~\citep{DeGroot1983,Niculescu-Mizil2005} and the \emph{expected calibration error metric}~\citep{Guo2017,Naeini2015}.

\begin{figure}[ht!!!]
	\centering
	\vspace{-5pt}
    \includegraphics[width=.65\columnwidth]{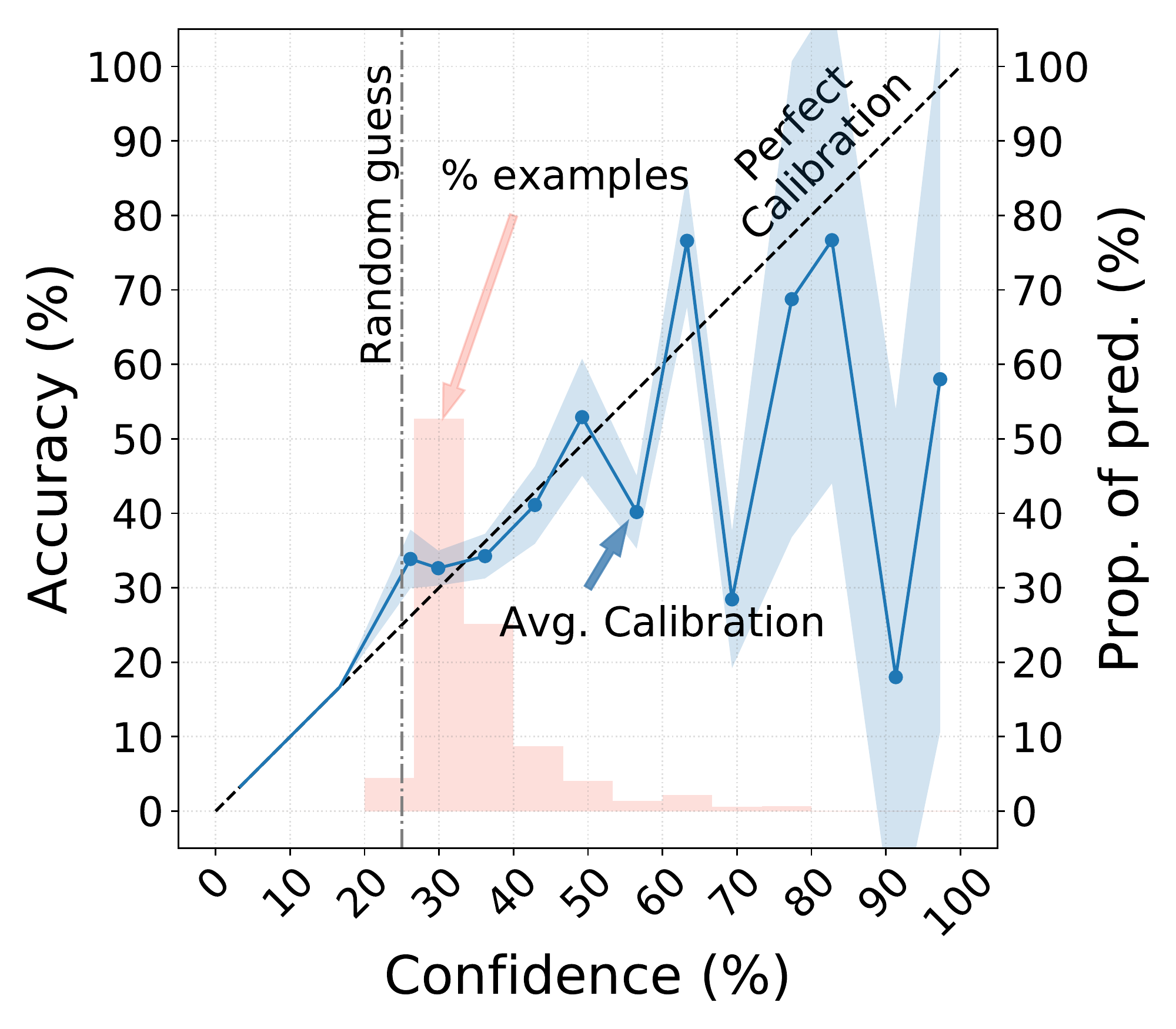}
	\vspace{-5pt}
	\caption{
	    \textbf{Reliability diagram}. (GCN on \textsc{Friendster}, with MC Dropout, after applying Temperature Scaling). 
	    The miscalibration of GCN is apparent, but ECE is just 4.29\% since nearly random predictions tend to be calibrated. \ECEOdds{} is 17.12\% and better matches how decision-makers are likely to interpret the plot. 
	    Average calibration (w/ std. deviation) is the blue line, histogram of predicted probabilities is given in the background; random guess and perfect calibration are given as reference.
	}
	\label{fig:reliability-diagram}
\end{figure}
\begin{table*}
    \centering
    \caption{\small 
    (Balanced Class Distribution) 
    Impact of shifted test distribution on evaluation of accuracy and calibration for \textsc{Friendster} is, in some cases, greater than that of applying a calibration method.
    We report Mean Accuracy, ECE, and \ECEOdds{}, w/ std. deviation from 10 bootstraps of test data, before and after applying Temperature Scaling. 
    For each GNN family we show the model with best (balanced) accuracy, full results in the \Appendix. 
    }
    \resizebox{\textwidth}{!}{
    \begin{tabular}{llrrrrrrrrrr}
    \toprule
        & & \multicolumn{5}{c}{Test distribution shift (imbalanced labels)} & \multicolumn{5}{c}{Test dist. = Train dist. (balanced labels)} \\
        \cmidrule(lr){3-7} \cmidrule(lr){8-12}
        & &
            \multicolumn{1}{c}{Accuracy (\%) \(\uparrow\)} &
            \multicolumn{2}{c}{ECE (\%) \(\downarrow\)} &
            \multicolumn{2}{c}{\ECEOdds (\%) \(\downarrow\)} &
            \multicolumn{1}{c}{Accuracy (\%) \(\uparrow\)} &
            \multicolumn{2}{c}{ECE (\%) \(\downarrow\)}  &
            \multicolumn{2}{c}{\ECEOdds (\%) \(\downarrow\)} \\
        \cmidrule(lr){3-3}  
        \cmidrule(lr){4-5}
        \cmidrule(lr){6-7}
        \cmidrule(lr){8-8} 
        \cmidrule(lr){9-10}
        \cmidrule(lr){11-12}
        Dataset & Model &
            \multicolumn{1}{c}{Original} &
            \multicolumn{1}{c}{Original} & \multicolumn{1}{c}{Temp. Scaling} &
            \multicolumn{1}{c}{Original} & \multicolumn{1}{c}{Temp. Scaling} &
            \multicolumn{1}{c}{Original} & 
            \multicolumn{1}{c}{Original} & \multicolumn{1}{c}{Temp. Scaling} &
            \multicolumn{1}{c}{Original} & \multicolumn{1}{c}{Temp. Scaling}\\
    \midrule
    \textsc{Friendster}    & GCN-MCD &  40.48 (00.69) &  11.04 (00.74) & 11.02 (00.64) & 26.36 (02.17) & 27.51 (02.62) & 35.09 (00.94) & 4.40 (01.21) & 4.29 (01.37) & 17.18 (04.40) & 17.12 (05.02) \\
                           & GAT     &  29.03 (00.40) &   9.39 (00.37) &  5.97 (00.39) & 48.89 (02.89) & 49.07 (02.51) & 32.40 (01.55) & 6.51 (01.88) & 3.63 (01.69) & 35.48 (04.78) & 36.50 (07.71) \\
                           & GIN-MCD &  23.91 (00.45) &   8.26 (00.33) &  7.81 (00.42) & 46.11 (06.00) & 46.48 (06.67) & 27.63 (01.13) & 3.18 (00.87) & 2.71 (00.81) & 30.83 (08.44) & 39.83 (12.66) \\
    \bottomrule
    \end{tabular}
    }
    \label{tab:balancing}
\end{table*}
\emph{Reliability Diagram.} Also called calibration curve \citep{DeGroot1983,Niculescu-Mizil2005}, this is a visual representation of the calibration error, across a range of confidence values within \([0, 1]\) (e.g. see \Cref{fig:reliability-diagram}).
To make this diagram, the predictions of the model are grouped in \(m\) bins, according to their confidence value.
For each bin \(B_k\), \(1 \leq k \leq m\), a point is drawn where the x-axis is the average confidence of the predictions in the bin:
\begin{align*}
\text{conf}(B_k) &= \frac{1}{|B_k|} \sum_{i \in B_k} \confidence[i],\\
\intertext{while the y-axis is their average accuracy:}
\text{acc}(B_k)  &= \frac{1}{|B_k|} \sum_{i \in B_k} \Ind[\target[i] = \predicted[i]],
\end{align*}
where \(|B_k|\) is the number of examples in the \(k\)-th bin and \(\Ind[\cdot]\) is the indicator function.

\emph{Expected Calibration Error (ECE).} This metric \citep{Naeini2015,Guo2017} is a single number that summarizes the calibration error.
ECE is the average of the gaps in the reliability diagram, weighted by the number of predictions in each bin, computed as
\begin{align*}
\text{ECE} &= \sum_{k=1}^m \; \frac{|B_k|}{n} \, \left|\text{acc}(B_k) - \text{conf}(B_k)\right|, \label{eq:ECE}
\end{align*}
where \(m\) is the number of bins and \(n\) the total number of examples.
ECE, however, can be small for models that make mostly random predictions regardless of the calibration of examples with high confidence, since random predictions tend to be calibrated. 
The model in \Cref{fig:reliability-diagram} has a low ECE (4.29\%) while decision-makers ---that care about making better-than-odds predictions--- would likely consider it an unreliable model.
In a better-than-odds prediction, being in the predicted class is more likely than not being in the predicted class.
To address this shortcoming of ECE, we propose \ECEOdds: an ECE computed only over examples with better-than-odds confidence (higher than 50\%). In the example of \Cref{fig:reliability-diagram}, \ECEOdds{} is 17.12\%, which better matches the confidence someone looking for better-than-odds predictions should have in the model.

%% file: miscalibration.tex
\section{Miscalibration of Graph Neural Networks}
\label{sec:miscalibration}

In this section we investigate the calibration of successful GNNs on a selection of datasets. 

\begin{table*}
    \centering
    \caption{
    (Calibration) 
    Avg. \ECEOdds and ECE --lower is better-- (w/ std. deviation) on 10 bootstraps of the test data, for best-performing (uncalibrated) model for each GNN family.
    MCD indicates MC Dropout. 
    For each row, we bold the best performing calibration method for ECE and \ECEOdds and the methods whose performance (with std. deviation) overlaps with it.
    For more results on other GNNs and datasets, see the \Appendix.
    }
    \resizebox{\textwidth}{!}{%
    \begin{tabular}{llrrrrrrrrr}
    \toprule
            &       & \multicolumn{2}{c}{Accuracy (\%) \(\uparrow\)} & 
                      \multicolumn{3}{c}{\(\text{ECE}\) (\%) \(\downarrow\)}  &  
                      \multicolumn{4}{c}{\ECEOdds (\%) \(\downarrow\)}  \\
    \cmidrule(lr){3-4} \cmidrule(lr){5-7} \cmidrule(lr){8-11}
    Dataset & Model & 
        \multicolumn{1}{c}{Original} & \multicolumn{1}{c}{Random} & 
        \multicolumn{1}{c}{Original} & \multicolumn{1}{c}{Isotonic Reg.} & \multicolumn{1}{c}{Temp. Scaling} & 
        \multicolumn{1}{c}{Original} & \multicolumn{1}{c}{Hist. Binning} & \multicolumn{1}{c}{Isotonic Reg.} & \multicolumn{1}{c}{Temp. Scaling} \\
    \midrule
    \textsc{Friendster} & GCN-MCD & 35.09 (00.94) & 25.00 & {\bf 4.40 (01.21)} & {\bf 6.43 (01.59)} & {\bf 4.29 (01.37)} & {\bf 17.18 (04.40)} & {    39.98 (00.18)} & {\bf 16.46 (05.34)} & {\bf 17.12 (05.02)} \\
                        & GAT     & 32.40 (01.55) & 25.00 & {\bf 6.51 (01.88)} & {\bf 4.13 (00.89)} & {\bf 3.63 (01.69)} & {\bf 35.48 (04.78)} & {    40.59 (00.16)} & {\bf 39.13 (04.83)} & {\bf 36.50 (07.71)} \\
                        & GIN-MCD & 27.63 (01.13) & 25.00 & {\bf 3.18 (00.87)} & {    7.00 (01.75)} & {\bf 2.71 (00.81)} & {\bf 30.83 (08.44)} & {    40.71 (00.17)} & {    46.04 (02.28)} & {\bf 39.83 (12.66)} \\
    \midrule
    \textsc{PubMed}     & GCN-MCD & 87.67 (00.50) & 33.33 & {    4.12 (00.45)} & {\bf 1.43 (00.38)} & {\bf 1.12 (00.24)} & {     4.05 (00.42)} & {     2.08 (00.43)} & {\bf  1.39 (00.37)} & {\bf 1.04 (00.23)} \\
                        & GAT-MCD & 87.03 (00.45) & 33.33 & {    5.07 (00.36)} & {\bf 2.01 (00.27)} & {\bf 1.64 (00.28)} & {     5.14 (00.37)} & {\bf  1.76 (00.26)} & {\bf  1.99 (00.26)} & {\bf 1.57 (00.32) } \\
                        & GIN     & 86.66 (00.42) & 33.33 & {    4.66 (00.37)} & {\bf 1.74 (00.31)} & {\bf 1.68 (00.36)} & {     4.62 (00.32)} & {\bf  1.80 (00.50)} & {\bf  1.69 (00.26)} & {\bf 1.56 (00.34) } \\
    \midrule
    \textsc{CiteSeer}   & GCN-MCD & 74.90 (01.32) & 16.67 & {   23.53 (01.29)} & {\bf 6.45 (01.49)} & {\bf 5.22 (01.06)} & {    24.23 (02.12)} & {\bf  5.46 (01.62)} & {\bf  5.59 (01.64)} & {\bf  5.27 (01.18)} \\
                        & GAT     & 73.57 (01.62) & 16.67 & {   29.85 (01.64)} & {\bf 7.45 (01.43)} & {\bf 6.82 (01.75)} & {    28.55 (01.80)} & {\bf  5.91 (01.43)} & {\bf  7.16 (01.62)} & {\bf  5.53 (01.33)} \\
                        & GIN-MCD & 62.88 (01.24) & 16.67 & {\bf 5.67 (01.51)} & {\bf 6.92 (01.31)} & {\bf 5.95 (01.19)} & {\bf  5.43 (01.62)} & {\bf  5.52 (01.70)} & {\bf  6.06 (01.99)} & {\bf  6.55 (01.93)} \\
    \bottomrule
    \end{tabular}
    }
    \label{tab:calibration-results}
\end{table*}

\begin{figure*}[t]
     \centering
     \begin{subfigure}[b]{0.245\textwidth}
         \centering
         \includegraphics[width=\textwidth]{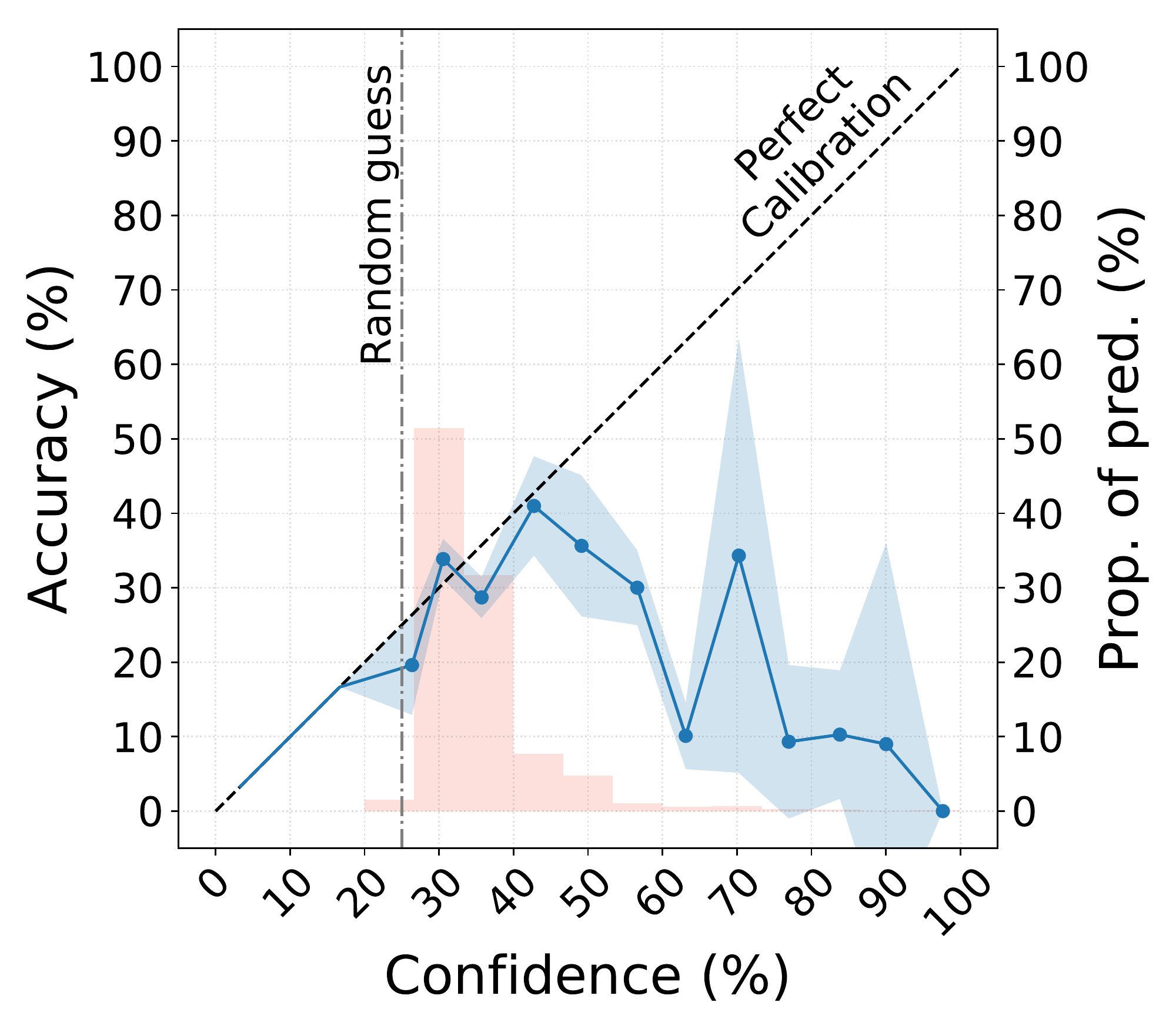}
         \caption{GAT on \textsc{Friendster}\\w/o calibrating}
         \label{fig:friendster-gat-before-ts}
     \end{subfigure}
     \begin{subfigure}[b]{0.245\textwidth}
         \centering
         \includegraphics[width=\textwidth]{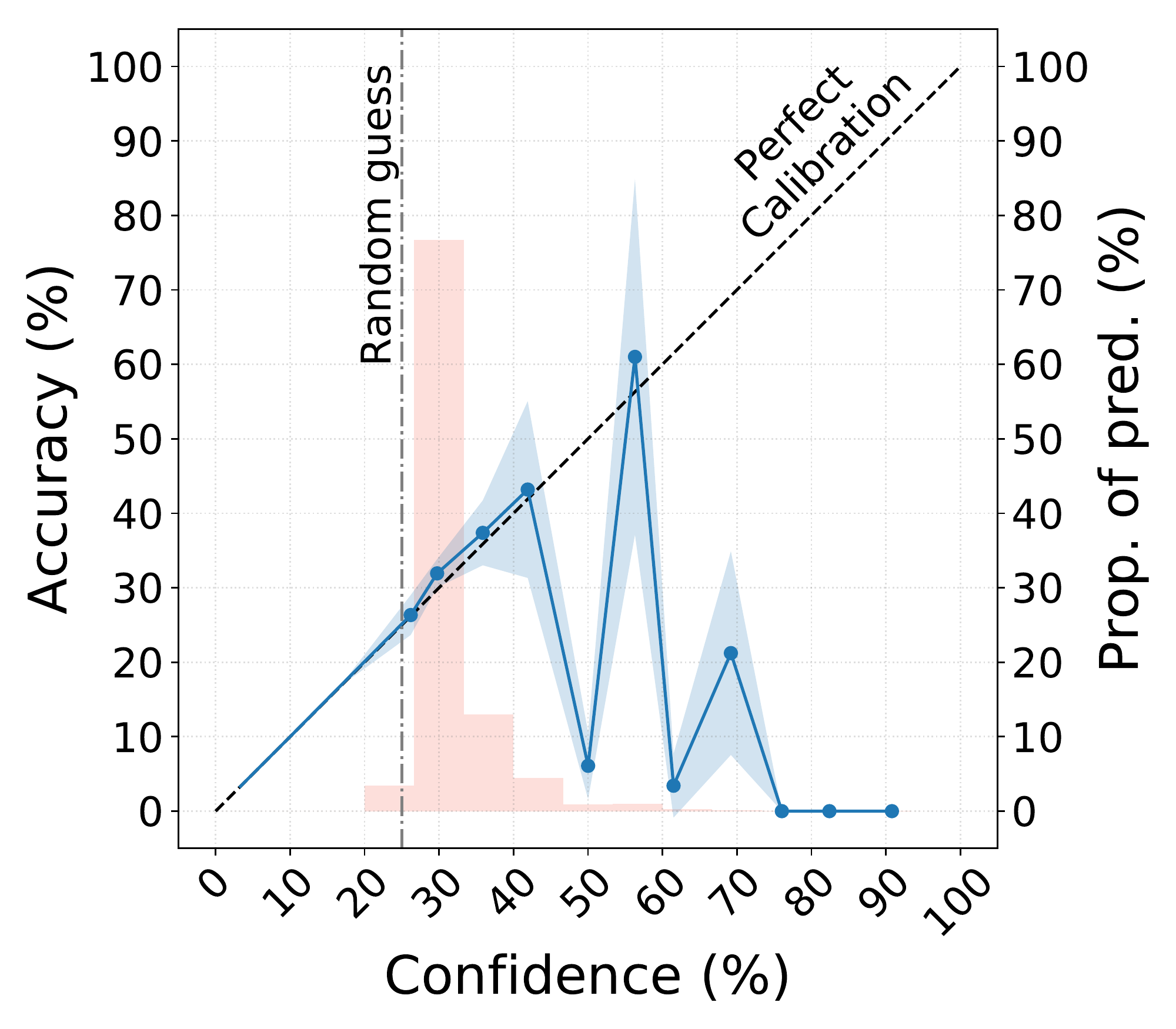}
         \caption{GAT on \textsc{Friendster}\\w/ Temp. Scaling}
         \label{fig:friendster-gat-after-ts}
     \end{subfigure}
     \hfill
     \begin{subfigure}[b]{0.245\textwidth}
         \centering
         \includegraphics[width=\textwidth]{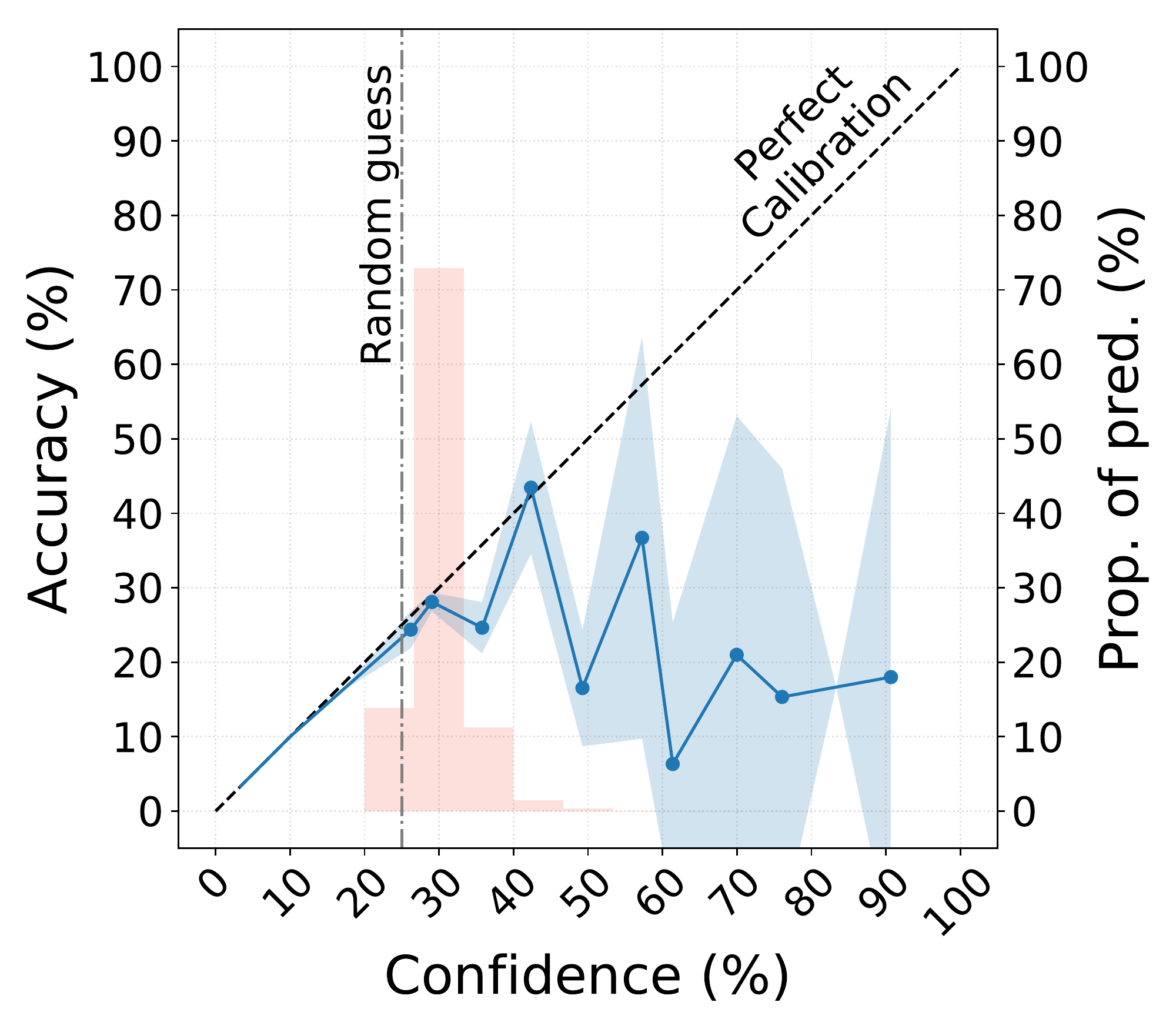}
         \caption{GIN-MCD on \textsc{Friendster}\\w/ Temp. Scaling}
         \label{fig:friendster-gin-ts}
     \end{subfigure}
     \hfill
     \begin{subfigure}[b]{0.245\textwidth}
         \centering
         \includegraphics[width=\textwidth]{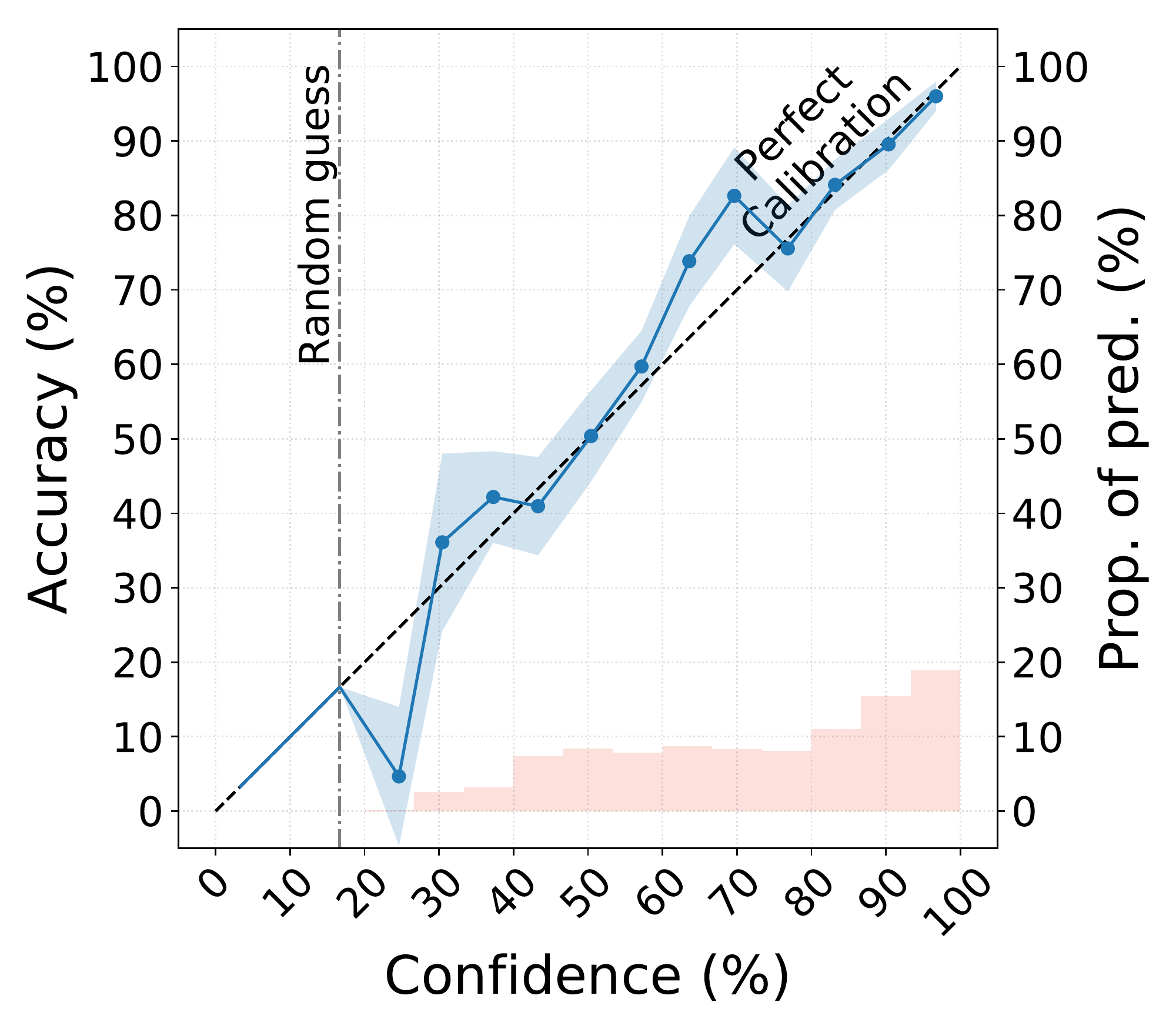}
         \caption{GCN-MCD on \textsc{Citeseer}\\w/ Temp. Scaling}
         \label{fig:citeseer-gcn-ts}
     \end{subfigure}
    \caption{
    (a) and (b) show GAT for \textsc{Friendster}, before and after applying Temperature Scaling. 
    (c) is GIN w/ MC Dropout for \textsc{Friendster}, after Temperature Scaling and (d) is GCN w/ MC Dropout for \textsc{CiteSeer}, after Temperature Scaling.}
    \label{fig:rc-temp-scaling}
\end{figure*}

\paragraph{Datasets and GNN models.}%
\label{sec:datasets}
We train GNNs for the task of node classification in the following graphs:
\textsc{Friendster}\footnote{\url{https://github.com/PurdueMINDS/GNNsMiscalibrated}} (social network);
\textsc{CiteSeer}, \textsc{PubMed} \citep{Sen2008} (citation networks).
Detailed description, as well as results for other graphs (\textsc{Cora}, \textsc{Amazon} and \textsc{Facebook}) can be found in the \Appendix.

We train the following GNNs: Graph Convolutional Networks (GCN)~\citep{Kipf2017}, Graph Attention Networks (GAT)~\citep{Velickovic2018} and Graph Isomorphism Network (GIN)~\citep{Xu2018}.
Our implementation uses the PyTorch-Geometric library \citep{Fey-Lenssen-2019}.
More details on hyperparameter search and optimization procedure in the \Appendix.

\paragraph{Existing calibration methods.} 
After training the GNNs, we apply techniques commonly used to improve the uncertainty quantification and the calibration of the model.
These include \textsc{MC Dropout} \citep{Gal2016}, \textsc{Histogram Binning} \citep{Zadrozny2001}, \textsc{Isotonic Regression} \citep{Zadrozny2002} and \textsc{Temperature Scaling} \citep{Guo2017}. 
More details of these methods and our experimental setup can be found in the \Appendix.

\paragraph{Calibrating for balanced classes.}%
In many datasets, it is common to face an imbalanced class distribution, which can pose challenges to learn meaningful models.
In our \textsc{Friendster} dataset, we observe a severe class imbalance, with the one class having 60\% of the nodes, while another class has less than 1\% of the nodes, which led to collapsed predictions towards a single (most prevalent) class, with all GNNs predicting at least 95\% of examples to a single class (or, in some cases, even predicting 100\% examples in a single class).
To overcome this, we force the class distribution to be balanced during the training, by weighting the loss function ---{\em i.e.}, upweighting the examples of the less prevalent classes--- such that all examples contribute equally.

When the test and train distribution are different, conclusions drawn from the evaluation can be misleading. 
Thus, in our case of a balanced-class loss function in training, it is paramount to evaluate the model with balanced class distributions in testing, by applying the same weighting scheme when computing the test metrics (Accuracy and ECE). 
As the results in \Cref{tab:balancing} show, using the proper test distribution when evaluating has an impact on both accuracy and calibration. 
In particular, for our hardest task (\textsc{Friendster}), we see that evaluating under a test distribution similar to that used for training has an impact on ECE greater than that of calibrating the trained model.
More details and results for other dataset can be found in the \Appendix.

\paragraph{Results.}%
In \Cref{tab:calibration-results} we present the results for the GNNs applied to \textsc{Friendster} ---our hardest task--- as well as for two common benchmark datasets. 
For tasks other than \textsc{Friendster}, existing methods are capable of improving calibration, with temperature scaling usually giving the best results. 
However, we also see that for harder tasks (\textsc{Friendster}), none of the existing calibration methods we tried were enough to fix it, particularly failing at the less frequent (but overconfident) predictions.
Moreover, we see that ECE does not capture miscalibration of such predictions, which might be essential for high-stakes decision making, while the proposed \ECEOdds presents itself as a useful metric.

As an example, we see that the ECE for GIN with MC-Dropout on \textsc{Friendster}, with Temperature Scaling, is less than half of the value of ECE for GCN with MC-Dropout and Temperature Scaling for \textsc{CiteSeer}, indicating a better calibrated model, while from a visual inspection of the reliability diagrams (\Cref{fig:friendster-gin-ts,fig:citeseer-gcn-ts}) we would make the opposite conclusion. The \ECEOdds metric captures the miscalibration that is apparent in the diagrams.

The diagrams in \Cref{fig:rc-temp-scaling} also show an important aspect of ECE: while applying Temperature Scaling to the GAT model in \textsc{Friendster} produces a model with lower ECE, the calibration of the more confident predictions actually gets worse (\Cref{fig:friendster-gat-before-ts,fig:friendster-gat-after-ts}).
As the Temperature Scaling method minimizes the negative log-likelihood, the region of close-to-random predictions (which has a larger fraction of examples) has higher impact on the loss (and ECE). 
What we observe is that those predictions become calibrated (which brings the ECE down), at the expense of making the calibration of the less frequent predictions (confidence above 50\%) worse, as those examples have a smaller impact in the loss.
In this case, the proposed \ECEOdds captures that those predictions remain miscalibrated.

%% file: conclusion.tex
\section{Conclusion}
In this work we empirically investigate the calibration of Graph Neural Networks (GNNs).
Our results show that for easier tasks all GNNs are reasonably calibrated, while  for harder tasks, such as our \textsc{Friendster} dataset, GNNs can be miscalibrated and existing calibration techniques are unable to calibrate them.
We also propose a new ECE-derived calibration metric.
Our results show the need to develop new methods to improve GNN calibration to increase their trustworthiness in high-stakes decision-making.

%% file: appendix.tex

\twocolumn[

\icmltitle{\Appendix for `Are Graph Neural Networks Miscalibrated?'}
\icmltitlerunning{\Appendix for `Are Graph Neural Networks Miscalibrated?'}




\vskip 0.3in
]

\appendix

\section{Datasets and GNN models}

In this section we give more detailed information on the datasets and models we used.


\subsection{Datasets}

The datasets we used are composed of citation networks, social networks and co-purchased goods. 
For all datasets, we randomly split the labeled nodes into three sets: training, validation and test, in the proportions described in \Cref{tab:datasets}.
A brief description of each dataset is given in the following paragraphs.

\paragraph{\textsc{Friendster}:} social network, where nodes represent users and edges represent friendship relationships. 
The features of the nodes include numerical features (e.g age, number of photos posted, etc) and categorical features (e.g. gender, college, music interests, etc), encoded as binary one-hot features, for a total of 644 features.
As an extra pre-processing step, we standardize all the features to have mean 0 and variance 1, across all nodes.
The predicted label is the relationship status of the user, which can be one of four values: Single, Married, In A Relationship or Domestic Partner. 
There are also nodes without label, which we use when computing the embeddings, but are outside of the set of labeled nodes we use to train (compute the loss) and evaluate the model.
The full graph has a largest connected component of more than 6 million nodes. 
However, due to the challenges of training GNNs on large graphs of this size, we obtained a sample of the larger graph, using the Forest Fire procedure \citep{Leskovec2006}. 
This smaller version of the graph contain 40K nodes, 25K of which are labeled.

\paragraph{\textsc{Facebook} \citep{Yang2017}:}
social network of Facebook users from Purdue university, where nodes represent users and edges represent friendship relationships. 
The feature of the nodes are: religious views, sex and whether the user's hometown is in Indiana.
In addition to those features, we also add the degree of the nodes, as one-hot encoding, bringing the total number of features to 85.
As an extra pre-processing step, we standardize all the features to have mean 0 and variance 1, across all nodes.
The predicted label is the political view.
The graph used is a subset of the entire graph used in \citet{Yang2017}, composed by all the users who represented all the features.

\paragraph{\textsc{Cora}, \textsc{CiteSeer}, \textsc{PubMed} \citep{Sen2008}:} 
citation networks, where the nodes represent papers and the edges represent a citation (undirected) between papers.
The features of the nodes are textual features (bag-of-words).
As in \citep{Kipf2017}, we normalize the features of each node, to have unitary norm.
The predicted label is the topic of the paper.
Note that, while some other works (e.g. \citet{Kipf2017}) employ a semi-supervised setting, using only a small fraction of nodes for training, we follow a supervised setting, where all nodes are used for either training, validation or testing.

\paragraph{\textsc{Cora-Full} \citep{Shchur2018}:} 
an extended version of \textsc{Cora}, with a larger number of nodes, features and labels. 
As before, nodes represent papers and edges represent a (undirected) citation between papers.
Node features are textual representations of the content and labels are the topic of the paper.

\paragraph{\textsc{Amazon-Computers}, \textsc{Amazon-Photo} \citep{Shchur2018}}: 
segments of the co-purchased graph from Amazon.
Nodes are goods, edges between nodes indicate they are frequently co-purchased.
Features are bag-of-words encoding of reviews of the product, while labels are given by the category.

\Cref{tab:datasets} gives detailed statistics of the datasets.

\begin{table*}[htb]
    \centering
    \begin{tabular}{lrrrrrrrr}
    \toprule
        Dataset & Classes & Features & Nodes & Edges & Edge density & Train & Validation & Test split \\
        \midrule
        \textsc{Friendster}         &  4 &  644 & 43880 & 145407 & 0.00015 &  15629 &  3126 &  6251\\
        \textsc{Facebook}           &  2 &   85 &  4556 &  23325 & 0.00225 &   2848 &   569 &  1139\\
        \textsc{Cora}               &  7 & 1433 &  2708 &   5278 & 0.00144 &   1208 &   500 &  1000\\
        \textsc{CiteSeer}           &  6 & 3703 &  3327 &   4552 & 0.00082 &   2080 &   416 &   831\\
        \textsc{PubMed}             &  3 &  500 & 19717 &  44324 & 0.00023 &  12324 &  2464 &  4929\\
        \textsc{Cora-Full}          & 70 & 8710 & 19793 &  63421 & 0.00032 &  12371 &  2474 &  4948\\
        \textsc{Amazon-Cmp.}   & 10 &  767 & 13752 & 245861 & 0.00260 &   8595 &  1719 &  3438\\
        \textsc{Amazon-Ph.}       &  8 &  745 &  7650 & 119081 & 0.00407 &   4782 &   956 &  1912\\
        \bottomrule
    \end{tabular}
    \caption{Dataset statistics and information on the splits used. The \emph{edge density} is the fraction of all possible edges that is present in the graph.}
    \label{tab:datasets}
\end{table*}


\subsection{GNN models}

In our experiments, we employed the Graph Convolutional Networks (GCN) from \citet{Kipf2017}, Graph Attention Networks (GAT) from \citet{Velickovic2018} and Graph Isomorphism Networks (GIN) from \cite{Xu2018}.
The models were implemented using the PyTorch-Geometric library \cite{Fey-Lenssen-2019}.
For GIN, we learn the parameter \(\epsilon\) and use a two-layer MLP with Batch Normalization \citep{Ioffe2015} ReLU activation \citep{Nair2010} at each layer.

\paragraph{MC-Dropout} \citep{Gal2016} is a simple way of improving the uncertainty estimations of a model with Dropout, by doing multiple forward passes, sampling a different dropout mask each time and averaging the results (instead of the mean approximation of using a single forward pass, without dropping neurons but multiplying them by the dropout rate).
While a useful (and simple) way of improving the conditional probability estimated by the model, this procedure does not targeted at enforcing a calibrated model, and a calibration method can be applied on top of it, to improve the model calibration.
In our experiments, we use 100 forward passes when applying MC Dropout. 
We use the suffix ``-MCD'' to denote when we apply MC Dropout with 100 forward passes to the trained GNN.

\subsection{Calibration Methods}

For the calibration methods, we employed three procedures which have been previously applied in the literature, which we describe here.
Similarly to what was done in \citet{Guo2017}, for Histogram Binning and Isotonic Regression, we train one version of the model for each class in a one-vs-all manner, which means that after the calibration, the estimated probabilities for one example need not sum to one across all classes and the predicted class might change, based on the transformed confidence values for each class (although we observed that this happen with very low frequency).

\paragraph{Histogram Binning} \citep{Zadrozny2001} is a simple method which groups the predictions into bins, according to their confidence values (similar to what is done for ECE and the reliability diagrams).
Then we build a mapping from the confidence range of each bin to the accuracy of the predictions of that bin, so that when a new prediction is made, we need only to see which bin it originally falls into and replace the confidence with the accuracy of that bin.
In our experiments we use 15 bins.

\paragraph{Isotonic Regression} \citep{Zadrozny2002} can be seen as a more general version of Histogram Binning, where the number of bins and their limits are jointly learned with the piece-wise (isotonic) regression on the accuracy. As with histogram binning, we fit one model for each class in one-vs-all encoding.
We use the Isotonic Regression implementation available in the \texttt{scikit-learn} Python package \citep{scikit-learn}.

\paragraph{Temperature Scaling} \citep{Guo2017} is a simple extension of Platt scaling \citep{Platt1999} to a multi-class setting, proposed by \citet{Guo2017}. 
A single scalar \emph{temperature} parameter is learned.
This temperature parameter (\(> 0\)) scales the logit (pre-softmax) values, which will alter the estimated probabilities, without changing the predicted class.
In their original paper, \citet{Guo2017} learn this parameter by optimizing the negative log-likelihood on validation data.
We tested using both the negative log-likelihood as well as the \emph{Brier Score} \citep{Brier1950} which is a proper scoring rule that can be decomposed into a calibration and refinement term \citep{Blattenberger1985}.
We implement it using the optimization routines from SciPy \citep{SciPy}.


\begin{table*}
    \centering
    \caption{
        Measurements of dispersion for the number of examples in each class. 
        \emph{Imbalance Ratio} is the ratio between the largest and smallest class; 
        \emph{Entropy} is Shannon's entropy; 
        \emph{Simpson Index} measures the probability that two independently and uniformly sampled examples are from the same class (in parenthesis is how larger the index is than if the dataset was balanced);
        and \emph{Eff. Number of Classes} is the inverse of the Simpson Index.
    }
    \begin{tabular}{lrrrrr}
        \toprule
            \multirow{1}{*}[-6pt]{Dataset} 
            & \multicolumn{1}{p{3cm}}{\small\centering Imbalance Ratio \\ Largest / Smallest class} 
            & \multirow{1}{*}[-6pt]{Entropy} 
            & \multicolumn{1}{p{2cm}}{\centering Simpson\\Index}
            & \multicolumn{1}{p{2cm}}{\small\centering Eff. Num. \\ of Classes} 
            & \multirow{1}{*}[-6pt]{Num. Classes} 
            \\
        \midrule
        \textsc{Friendster}  & 91.62\(\times\) & 1.20 & 0.35 ( 1.38\(\times\) ) &  2.9 &  4 \\
        \textsc{Cora}        &  4.54\(\times\) & 1.83 & 0.18 ( 1.26\(\times\) ) &  5.6 &  7 \\
        \textsc{PubMed}      &  1.92\(\times\) & 1.06 & 0.36 ( 1.07\(\times\) ) &  2.8 &  3 \\
        \textsc{CiteSeer}    &  2.82\(\times\) & 1.75 & 0.18 ( 1.07\(\times\) ) &  5.6 &  6 \\
        \textsc{Cora-Full}   & 61.87\(\times\) & 4.00 & 0.02 ( 1.52\(\times\) ) & 46.1 & 70 \\
        \textsc{Facebook}    &  1.91\(\times\) & 0.64 & 0.55 ( 1.10\(\times\) ) &  1.8 &  2 \\
        \textsc{Amazon-Cmp.} & 17.73\(\times\) & 1.87 & 0.21 ( 2.08\(\times\) ) &  4.8 & 10 \\
        \textsc{Amazon-Ph.}  &  5.86\(\times\) & 1.93 & 0.16 ( 1.26\(\times\) ) &  6.1 &  8 \\
    \bottomrule
    \end{tabular}
    \label{tab:dispersion}
\end{table*}

\section{Experimental setup and hyperparameters}

For all GNNs, we tested 2, 3, and 4 layers. The number of neurons in the hidden layers was chosen from \(\{64, 128, 512\}\) for GCN and GIN, \(\{16, 32, 50\}\) for GAT (with 8 heads at each layer which are concatenated for intermediary layers and average for last layer, as in the original paper). 
For all models we used Dropout in the final fully connected layers, with the probability of zeroing a neuron chosen from \(\{0.1, 0.5, 0.8\}\). 
The strength of weight decay was chosen from \(\{0, 5 \times 10^{-4}\}\).
For GIN, we learn the parameter \(\epsilon\).
We trained all models to minimize the \emph{negative log-likelihood} of the training data using full-batch gradient descent and the Adam optimizer \citep{Kingma2014} with learning rate chosen from \(\{10^{-2}, 10^{-3}\}\) and default values of betas (\(0.9\) and \(0.999\)).
We trained our models for 200 epochs, stopping early if the validation performance does not improve after 50 consecutive epochs.
We also decay the learning rate by a factor of 2 after each 75 epochs.

For each GNN family, we select the best model (hyperparameters) as the one which achieves best performance in the validation data.
We tested using either accuracy or loss as the performance metric for early stopping and model selection.
As this choice does not seem to have too much impact on the calibration of the model, we decided to use accuracy for final results presented in the paper, as this choice yields models with slightly better accuracy.
In \Cref{sec:results-appendix} we also present results using the validation loss as the metric for model selection and early stopping, for comparison. 
All models were trained using a NVidia Titan V GPU, in a host with 2.00GHz Intel(R) Xeon(R) CPU E5-2660 v4 processor and 256 GiB of RAM. 

After training and selecting the best model for each GNN family, we evaluate the model (accuracy and calibration) on the test data.
We compare using MC Dropout (MCD suffix on the tables) or just the regular dropout.
For the evaluation, we perform 10 bootstraps of the test data and present our results as the mean and standard deviation over the bootstrap samples.

Since the calibration methods we employ are very simple and do not have hyperparameters to tune, we use the validation data to train them.
After training them, we evaluate them on the test data, using the same procedure described before.
For both the ECE metric and the Reliability Diagrams, we use 15 bins, which is the same number of bins used by \citet{Guo2017}.
We also use the same number of bins for histogram binning.


\section{Results}
\label{sec:results-appendix}

\begin{figure*}[ht!!]
    \centering
    \includegraphics[width=0.9\textwidth]{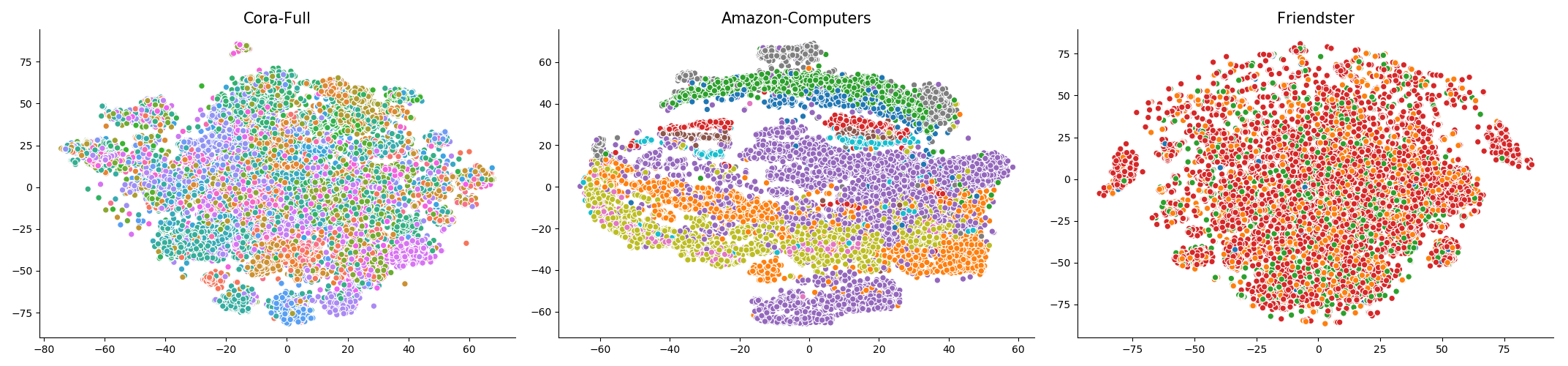}
    \caption{t-SNE embedding, computed from features of nodes, for the three datasets with largest size ratio between largest and smallest class. 
    Colors represent the true class of the nodes. For Friendster there is no clear clustering effect as we observe for other datasets.}
    \label{fig:tsne}
    \vspace{-8pt}
\end{figure*}


\subsection{Need for balancing test distribution for \textsc{Friendster}}

The class distribution in our \textsc{Friendster} dataset is severely imbalanced. 
The unlabeled nodes constitute 43\% of the nodes in the graph (while not used to compute the loss or the evaluation, they are used in the neighborhood of other nodes when computing their embeddings).
Of the remaining 25K labeled nodes, 66.5\% belong to the class \emph{Single}, 14.3 \% to the class \emph{Married}, 18.4 \% to the class \emph{In A Relationship} and 0.8 \% to the class \emph{Domestic Partner}.

This severe class imbalance leads to challenges during the training, such as collapsing the predictions towards the most prevalent class.
In our experiments, when training the GNNs without weighting the loss, all three models heavily biases their predictions towards the prevalent class, with GCN predicting 95.5\% of the test nodes as \emph{Single}, GAT predicting 96.2\%, and GIN 98.7\%.
For some configurations of hyperparameters, we even observed that the GNNs predicted all test nodes as \emph{Single} (we see this for all three GNNs).
These models achieve around 66\% accuracy in the test, which is the proportion of \emph{Single} nodes in the test data.

If we want our models to learn something more useful than just predicting any node as \emph{Single}, we need to deal with the class imbalance.
A common way of remedying this issue is to weight the loss function, such that each example contributes the same towards the loss, regardless of its class.
With this approach, we are adding and extra assumption to our model, that the distribution of the class labels is balanced.

It is important, however, to maintain this assumption when evaluating our models as well.
If we balance the classes only during training, but keep the test data unbalanced, we are evaluating our model under a distribution different from that for which it was trained. 
To avoid facing the challenges incurred by this type of domain adaptation, we must also evaluate our model with a balanced distribution, which can also be achieved by simply weighting the evaluation metrics in such a way that every example contributes the same, regardless of its class.

In \Cref{tab:dispersion} we present measurements of diversity of the distribution of class labels on each dataset. 
We compute: (1) the ratio between the size of the largest class and the size of the smallest class;
(2) entropy, as \(- \sum_c p_c \log p_c\), where \(p_c\) is the proportion of examples in class \(c\);
(3) Simpson's index \citep{Simpson1949}, given by \(\sum_c p_c^2\), which can be interpreted as the probability that two uniformly and independently sampled examples come from the same class, as well as how much larger this index is than if it was computed in a balanced dataset;
(4) Effective Number of Classes \citep{Laakso1979}, given by the inverse of Simpson's index.
As can be seen in the table, while Friendster has the most extreme imbalance when comparing the largest and smallest classes, other datasets also present some level of class imbalance.
Faced with these challenges, we decided for balancing the classes in all of our models, even for the other datasets, where the class imbalance is less severe.
We also employ the same balancing strategy when training the calibration methods and computing our evaluation metrics.

For the sake of comparison, we present here in the \Appendix the results when the model was trained with balanced classes but evaluated with imbalanced classes as well.
While the full results are presented in the next section, we also show a summary in \Cref{tab:balancing-appendix}, comparing the accuracy and ECE under both scenarios.
One interesting observation is how in some cases, such as for our harder \textsc{Friendster} dataset, simply evaluating under the balanced distribution (by weighting the metrics), as it was trained for, had a greater impact in ECE than applying a calibration method such as Temperature Scaling, but still evaluating under an imbalanced test distribution.

\begin{table*}
    \centering
    \resizebox{\textwidth}{!}{
    \begin{tabular}{llrrrrrr}
    \toprule
        & & \multicolumn{3}{c}{w/o Balancing} & \multicolumn{3}{c}{w/ Balancing} \\
        \cmidrule(lr){3-5} \cmidrule(lr){6-8}
        & &
            \multicolumn{1}{c}{Accuracy (\%) \(\uparrow\)} &
            \multicolumn{2}{c}{ECE (\%) \(\downarrow\)} &
            \multicolumn{1}{c}{Accuracy (\%) \(\uparrow\)} &
            \multicolumn{2}{c}{ECE (\%) \(\downarrow\)} \\
        \cmidrule(lr){3-3}  \cmidrule(lr){4-5}  \cmidrule(lr){6-6}  \cmidrule(lr){7-8}
        Dataset & Model &
            \multicolumn{1}{c}{Original} & \multicolumn{1}{c}{Original} & \multicolumn{1}{c}{Temp. Scaling} &
            \multicolumn{1}{c}{Original} & \multicolumn{1}{c}{Original} & \multicolumn{1}{c}{Temp. Scaling}\\
    \midrule
    \textsc{Friendster}    & GCN     &  39.70 (00.56) &  10.40 (00.78) & 10.90 (00.49) &  35.08 (02.32) &   4.08 (01.13) & 4.32 (01.27) \\
                          & GCN-MCD &  40.48 (00.69) &  11.04 (00.74) & 11.02 (00.64) &  35.09 (00.94) &   4.40 (01.21) & 4.29 (01.37) \\
                          & GAT     &  29.03 (00.40) &   9.39 (00.37) &  5.97 (00.39) &  32.40 (01.55) &   6.51 (01.88) & 3.63 (01.69) \\
                          & GAT-MCD &  28.53 (00.59) &   8.23 (00.54) &  5.22 (00.53) &  31.24 (01.63) &   5.60 (01.64) & 3.56 (01.43) \\
                          & GIN     &  23.84 (00.54) &   7.90 (00.45) &  7.68 (00.58) &  26.74 (00.84) &   3.43 (00.79) & 3.66 (00.64) \\
                          & GIN-MCD &  23.91 (00.45) &   8.26 (00.33) &  7.81 (00.42) &  27.63 (01.13) &   3.18 (00.87) & 2.71 (00.81) \\
    \midrule
    \textsc{Facebook}      & GCN     &  69.07 (01.49) &   4.93 (00.84) &  4.88 (01.50) &  70.15 (01.44) &   6.16 (01.55) & 5.49 (00.87) \\
                          & GCN-MCD &  71.44 (00.91) &   6.98 (00.84) &  4.67 (01.41) &  69.49 (01.08) &   5.31 (00.80) & 4.55 (01.08) \\
                          & GAT     &  67.45 (01.64) &   5.58 (01.72) &  4.21 (01.10) &  67.19 (02.08) &   5.92 (01.57) & 4.02 (00.76) \\
                          & GAT-MCD &  68.47 (01.27) &   5.94 (00.41) &  6.23 (00.98) &  68.62 (01.23) &   6.28 (00.93) & 5.75 (01.26) \\
                          & GIN     &  66.11 (01.97) &   5.81 (01.87) &  4.45 (00.89) &  65.11 (01.24) &   5.61 (00.79) & 5.65 (01.48) \\
                          & GIN-MCD &  65.56 (01.58) &   4.18 (01.16) &  4.76 (00.98) &  65.64 (01.68) &   4.78 (00.98) & 5.48 (01.58) \\
    \midrule
    \textsc{Cora}          & GCN     &  86.03 (01.25) &   4.76 (01.07) &  4.56 (01.00) &  86.00 (01.25) &   5.31 (01.08) & 5.53 (01.05) \\
                          & GCN-MCD &  86.06 (00.56) &  10.87 (00.64) &  4.23 (00.69) &  86.00 (01.02) &  10.23 (00.80) & 4.69 (00.79) \\
                          & GAT     &  85.29 (00.88) &  47.81 (00.80) &  4.30 (00.55) &  86.31 (01.07) &  48.01 (01.00) & 4.56 (00.88) \\
                          & GAT-MCD &  84.79 (01.01) &  47.15 (00.85) &  4.68 (00.79) &  85.30 (01.07) &  46.96 (01.22) & 5.01 (00.76) \\
                          & GIN     &  83.35 (01.14) &   9.65 (01.17) &  5.75 (00.99) &  84.61 (00.80) &   8.76 (00.72) & 6.03 (01.20) \\
                          & GIN-MCD &  83.77 (00.85) &   3.81 (00.62) &  4.90 (01.36) &  83.44 (01.49) &   5.18 (00.77) & 5.28 (01.21) \\
    \midrule
    \textsc{CiteSeer}      & GCN     &  77.21 (01.05) &  19.71 (00.92) &  5.56 (01.45) &  74.39 (01.76) &  17.71 (01.79) & 5.30 (00.85) \\
                          & GCN-MCD &  77.32 (01.48) &  25.27 (01.10) &  4.47 (01.12) &  74.90 (01.32) &  23.53 (01.29) & 5.84 (01.14) \\
                          & GAT     &  75.99 (01.11) &  30.68 (01.13) &  5.69 (00.87) &  73.57 (01.62) &  29.85 (01.64) & 7.11 (01.14) \\
                          & GAT-MCD &  76.49 (01.22) &  31.85 (01.23) &  6.86 (01.00) &  72.05 (01.82) &  29.37 (01.91) & 5.84 (01.38) \\
                          & GIN     &  67.00 (00.83) &   8.38 (00.77) &  6.07 (00.70) &  62.85 (01.70) &  11.75 (01.57) & 6.90 (01.24) \\
                          & GIN-MCD &  66.87 (01.47) &   7.63 (01.22) &  7.46 (01.44) &  62.88 (01.24) &   5.67 (01.51) & 6.28 (00.56) \\
    \midrule
    \textsc{PubMed}        & GCN     &  87.31 (00.57) &   1.53 (00.34) &  1.31 (00.37) &  87.63 (00.42) &   1.77 (00.30) & 1.25 (00.21) \\
                          & GCN-MCD &  87.51 (00.48) &   4.07 (00.36) &  1.38 (00.46) &  87.67 (00.50) &   4.12 (00.45) & 1.35 (00.20) \\
                          & GAT     &  86.71 (00.35) &   3.25 (00.37) &  1.50 (00.26) &  86.92 (00.50) &   3.42 (00.49) & 1.62 (00.26) \\
                          & GAT-MCD &  86.53 (00.25) &   4.85 (00.27) &  1.25 (00.32) &  87.03 (00.45) &   5.07 (00.36) & 1.64 (00.28) \\
                          & GIN     &  86.93 (00.40) &   4.66 (00.41) &  1.79 (00.34) &  86.66 (00.42) &   4.66 (00.37) & 1.68 (00.36) \\
                          & GIN-MCD &  86.84 (00.21) &   3.53 (00.26) &  1.86 (00.29) &  86.30 (00.59) &   3.85 (00.36) & 2.21 (00.26) \\
    \midrule
    \textsc{Amazon-Cmp.}   & GCN     &  87.97 (00.57) &   5.29 (00.52) &  2.90 (00.51) &  91.51 (00.54) &   3.77 (00.36) & 1.98 (00.48) \\
                          & GCN-MCD &  88.18 (00.47) &   5.56 (00.48) &  2.65 (00.47) &  91.51 (00.60) &   3.76 (00.59) & 1.81 (00.45) \\
                          & GAT     &  91.03 (00.48) &   2.15 (00.41) &  2.17 (00.32) &  93.15 (00.46) &   1.80 (00.33) & 1.94 (00.44) \\
                          & GAT-MCD &  90.84 (00.43) &   2.03 (00.40) &  2.58 (00.43) &  93.20 (00.26) &   2.26 (00.41) & 1.68 (00.35) \\
                          & GIN     &  88.66 (00.45) &   3.69 (00.58) &  3.67 (00.50) &  92.11 (00.53) &   2.85 (00.37) & 2.35 (00.44) \\
                          & GIN-MCD &  88.66 (00.42) &   7.86 (00.43) &  2.61 (00.29) &  92.08 (00.53) &   7.28 (00.46) & 2.27 (00.61) \\
    \midrule
    \textsc{Amazon-Ph.}    & GCN     &  93.06 (00.36) &   1.50 (00.34) &  1.92 (00.49) &  93.61 (00.69) &   1.71 (00.32) & 2.23 (00.44) \\
                          & GCN-MCD &  93.47 (00.37) &   1.74 (00.28) &  1.73 (00.23) &  93.68 (00.86) &   1.74 (00.34) & 2.19 (00.37) \\
                          & GAT     &  93.41 (00.60) &   2.84 (00.58) &  1.67 (00.42) &  93.38 (00.77) &   3.13 (00.53) & 2.20 (00.51) \\
                          & GAT-MCD &  93.11 (00.49) &   2.44 (00.42) &  1.89 (00.25) &  93.04 (00.47) &   2.54 (00.33) & 1.90 (00.39) \\
                          & GIN     &  92.51 (00.63) &   2.89 (00.83) &  2.51 (00.45) &  93.96 (00.59) &   2.74 (00.36) & 2.72 (00.63) \\
                          & GIN-MCD &  92.68 (00.53) &   2.32 (00.33) &  2.27 (00.41) &  93.87 (00.61) &   2.49 (00.55) & 2.74 (00.56) \\
    \bottomrule
    \end{tabular}
    }
    \caption{(Balancing Test Metrics) Mean Accuracy and ECE (and standard deviation) for original and calibrated model, with and without balancing the metrics according to the test distribution. 
    For each dataset we only show the GNN with best (balanced) accuracy.
    All models were trained with balanced loss.
    }
    \label{tab:balancing-appendix}
\end{table*}


\subsection{Misclassification on Friendster}

Among the datasets we tested, Friendster is the hardest task.
While for other tasks we achieve more than 90\% accuracy, all models tested on Friendster achieve accuracy levels closer to that of random predictions.

In \Cref{fig:tsne} we plot the nodes of the three datasets with largest ratio between largest and smallest class. 
The positions are computed from the node's features, using a Fast Fourier Transform-accelerated Interpolation-based implementation of t-SNE \citep{Linderman2019,maaten2008}.
The colors are the true labels.

As can be seen in the figure, for datasets other than Friendster, the features of the nodes have sufficient signal to produce a reasonable embedding of the nodes, while for Friendster, the embedding based on the features alone carries no information about the classes.
This could indicate that, for other tasks, the features of the nodes have a stronger signal when compared with the Friendster dataset.

We also investigated the examples misclassified for the GCN model on Friendster and found that the nodes with label ``In A Relationship'' have the highest misclassification rate (93\% misclassified).
Of the misclassified examples of this class, 47.3 \% are assigned the label ``Single'' and 46.4 \% are assigned as either ``Domestic Partner'' (14.1 \%) or ``Married'' (32.3\%).
Further investigation on why the performance is so poor on Friendster is left as future work.


\subsection{Calibration Results}

\begin{figure}[b!!]
    \centering
    \includegraphics[width=0.9\columnwidth]{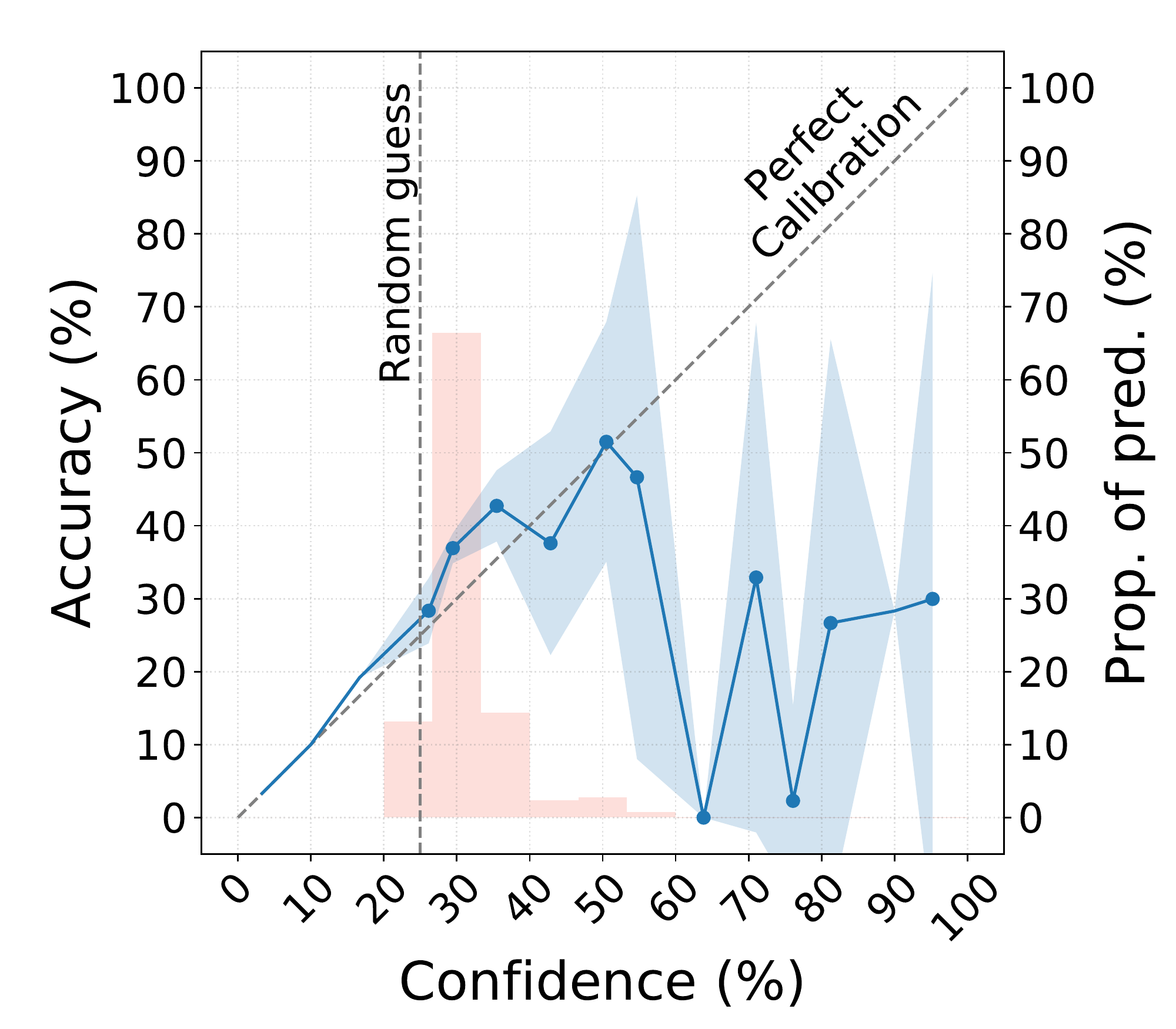}
    \caption{Logistic Regression with L1 regularization for Friendster, after applying Temperature Scaling. ECE is 7.5\% and \ECEOdds{} of 28.2\%.}
    \label{fig:logistic}
\end{figure}

In \Cref{tab:results-ece-accuracy-not-balanced,tab:results-ece-accuracy-balanced,tab:results-ece-loss-not-balanced,tab:results-ece-loss-balanced} we present the ECE results for all GNNs in all datasets, with and without balancing the metrics to correct for class imbalance, using accuracy or loss for early stopping / model selection.
The \ECEOdds results for the same models are present in the corresponding \Cref{tab:results-ece-50-accuracy-not-balanced,tab:results-ece-50-accuracy-balanced,tab:results-ece-50-loss-not-balanced,tab:results-ece-50-loss-balanced}.

We also investigate the calibration of a Logistic Regression model (with L1 regularization) for Friendster. 
This model obtains accuracy levels similar to GCN with MC Dropout, but still miscalibrated. 
In \Cref{fig:logistic} we present the reliability diagram, after applying Temperature Scaling.
Even when applying a non-i.i.d. model, the state-of-the-art calibration method is not capable of solving the problem, which reinforces the need to develop calibration methods that take into account the non-i.i.d. nature of the data as well.


\begin{table*}
\centering
\resizebox{\textwidth}{!}{
\begin{tabular}{llrrrrrrr}

\toprule
\multicolumn{9}{c}{Early Stopping and Model Selection using: Balanced Validation Accuracy, Test Metrics: Not Balanced} \\
\midrule
        &       & \multicolumn{2}{c}{ Accuracy (\%) }     &                                 \multicolumn{5}{c}{\(\text{ECE}\) (\%)} \\
\cmidrule(lr){3-4} \cmidrule(lr){5-9}
Dataset & Model & \multicolumn{1}{c}{Original} & \multicolumn{1}{c}{Random} & \multicolumn{1}{c}{Original}  & \multicolumn{1}{c}{Hist. Bin.}  & \multicolumn{1}{c}{Isotonic} & \multicolumn{1}{c}{TS w/ NLL} & \multicolumn{1}{c}{TS w/ Brier} \\
\midrule
\textsc{Friendster}    & GCN     &  39.70 (00.56) &    25.00 &  10.40 (00.78) &   3.79 (00.47) &   6.69 (00.80) &  10.90 (00.49) &  10.68 (00.79)  \\
                       & GCN-MCD &  40.48 (00.69) &    25.00 &  11.04 (00.74) &   4.67 (00.55) &   8.84 (00.52) &  11.02 (00.64) &  10.98 (00.44)  \\
                       & GAT     &  29.03 (00.40) &    25.00 &   9.39 (00.37) &   2.73 (00.45) &  15.22 (00.50) &   5.97 (00.39) &   6.89 (00.62)  \\
                       & GAT-MCD &  28.53 (00.59) &    25.00 &   8.23 (00.54) &   3.29 (00.69) &  17.05 (00.62) &   5.22 (00.53) &   6.39 (00.43)  \\
                       & GIN     &  23.84 (00.54) &    25.00 &   7.90 (00.45) &   1.31 (00.60) &  10.70 (00.58) &   7.68 (00.58) &   8.02 (00.66)  \\
                       & GIN-MCD &  23.91 (00.45) &    25.00 &   8.26 (00.33) &   1.40 (00.44) &   8.54 (00.56) &   7.81 (00.42) &   7.99 (00.56)  \\
\midrule
\textsc{Cora}          & GCN     &  86.03 (01.25) &    14.29 &   4.76 (01.07) &   5.87 (00.92) &   6.99 (00.83) &   4.56 (01.00) &   6.51 (00.90)  \\
                       & GCN-MCD &  86.06 (00.56) &    14.29 &  10.87 (00.64) &   4.93 (00.49) &   5.22 (01.19) &   4.23 (00.69) &   5.19 (00.91)  \\
                       & GAT     &  85.29 (00.88) &    14.29 &  47.81 (00.80) &   6.19 (01.26) &   3.75 (01.09) &   4.30 (00.55) &   4.17 (00.93)  \\
                       & GAT-MCD &  84.79 (01.01) &    14.29 &  47.15 (00.85) &   4.18 (01.00) &   3.20 (00.73) &   4.68 (00.79) &   5.01 (00.84)  \\
                       & GIN     &  83.35 (01.14) &    14.29 &   9.65 (01.17) &   5.80 (01.26) &   5.55 (01.05) &   5.75 (00.99) &   6.55 (01.25)  \\
                       & GIN-MCD &  83.77 (00.85) &    14.29 &   3.81 (00.62) &   6.32 (00.97) &   5.21 (00.76) &   4.90 (01.36) &   4.53 (00.82)  \\
\midrule
\textsc{PubMed}        & GCN     &  87.31 (00.57) &    33.33 &   1.53 (00.34) &   1.63 (00.36) &   1.90 (00.34) &   1.31 (00.37) &   1.20 (00.35)  \\
                       & GCN-MCD &  87.51 (00.48) &    33.33 &   4.07 (00.36) &   2.01 (00.22) &   1.87 (00.26) &   1.38 (00.46) &   1.38 (00.48)  \\
                       & GAT     &  86.71 (00.35) &    33.33 &   3.25 (00.37) &   1.69 (00.41) &   1.49 (00.25) &   1.50 (00.26) &   1.41 (00.30)  \\
                       & GAT-MCD &  86.53 (00.25) &    33.33 &   4.85 (00.27) &   1.61 (00.26) &   1.76 (00.34) &   1.25 (00.32) &   1.42 (00.22)  \\
                       & GIN     &  86.93 (00.40) &    33.33 &   4.66 (00.41) &   1.61 (00.44) &   1.80 (00.22) &   1.79 (00.34) &   1.71 (00.34)  \\
                       & GIN-MCD &  86.84 (00.21) &    33.33 &   3.53 (00.26) &   1.45 (00.31) &   2.33 (00.34) &   1.86 (00.29) &   2.29 (00.26)  \\
\midrule
\textsc{CiteSeer}      & GCN     &  77.21 (01.05) &    16.67 &  19.71 (00.92) &   4.67 (01.33) &   5.09 (01.13) &   5.56 (01.45) &   5.92 (01.01)  \\
                       & GCN-MCD &  77.32 (01.48) &    16.67 &  25.27 (01.10) &   4.45 (00.84) &   5.22 (00.76) &   4.47 (01.12) &   4.65 (00.90)  \\
                       & GAT     &  75.99 (01.11) &    16.67 &  30.68 (01.13) &   5.46 (00.91) &   7.42 (01.20) &   5.69 (00.87) &   5.16 (00.75)  \\
                       & GAT-MCD &  76.49 (01.22) &    16.67 &  31.85 (01.23) &   6.29 (00.69) &   4.74 (00.53) &   6.86 (01.00) &   5.60 (01.30)  \\
                       & GIN     &  67.00 (00.83) &    16.67 &   8.38 (00.77) &   5.99 (01.11) &   6.08 (00.78) &   6.07 (00.70) &   7.23 (01.29)  \\
                       & GIN-MCD &  66.87 (01.47) &    16.67 &   7.63 (01.22) &   5.93 (01.36) &   7.46 (00.99) &   7.46 (01.44) &   7.01 (01.40)  \\
\midrule
\textsc{Cora-Full}     & GCN     &  70.34 (00.61) &     1.43 &   5.79 (00.59) &   5.98 (00.61) &   5.68 (00.43) &   5.46 (00.54) &   4.99 (00.31)  \\
                       & GCN-MCD &  70.31 (00.55) &     1.43 &   6.83 (00.41) &   4.25 (00.47) &   3.62 (00.46) &   3.78 (00.47) &   3.49 (00.56)  \\
                       & GAT     &  69.44 (00.51) &     1.43 &   3.58 (00.48) &   4.63 (00.41) &   4.26 (00.37) &   4.36 (00.59) &   3.73 (00.49)  \\
                       & GAT-MCD &  69.90 (00.54) &     1.43 &   3.48 (00.54) &   4.00 (00.32) &   3.94 (00.40) &   3.59 (00.75) &   3.21 (00.35)  \\
                       & GIN     &  68.84 (00.45) &     1.43 &  13.48 (00.75) &   5.43 (00.62) &   5.20 (00.44) &  11.32 (00.42) &   9.41 (00.20)  \\
                       & GIN-MCD &  68.94 (00.77) &     1.43 &  31.76 (00.60) &   5.77 (00.55) &   3.46 (00.49) &   6.56 (00.43) &   6.13 (00.74)  \\
\midrule
\textsc{Facebook}      & GCN     &  69.07 (01.49) &    50.00 &   4.93 (00.84) &   5.21 (01.01) &   9.81 (01.31) &   4.88 (01.50) &   5.15 (01.08)  \\
                       & GCN-MCD &  71.44 (00.91) &    50.00 &   6.98 (00.84) &   5.59 (00.70) &   8.67 (00.95) &   4.67 (01.41) &   5.20 (00.95)  \\
                       & GAT     &  67.45 (01.64) &    50.00 &   5.58 (01.72) &   5.36 (00.91) &  17.37 (00.73) &   4.21 (01.10) &   4.75 (00.94)  \\
                       & GAT-MCD &  68.47 (01.27) &    50.00 &   5.94 (00.41) &   6.31 (01.42) &  11.83 (01.35) &   6.23 (00.98) &   5.27 (01.12)  \\
                       & GIN     &  66.11 (01.97) &    50.00 &   5.81 (01.87) &   3.73 (01.02) &   8.92 (01.02) &   4.45 (00.89) &   5.52 (01.26)  \\
                       & GIN-MCD &  65.56 (01.58) &    50.00 &   4.18 (01.16) &   4.84 (00.85) &   8.10 (01.71) &   4.76 (00.98) &   4.98 (01.13)  \\
\midrule
\textsc{Amazon-Cmp.}   & GCN     &  87.97 (00.57) &    10.00 &   5.29 (00.52) &   1.78 (00.41) &   3.76 (00.24) &   2.90 (00.51) &   3.49 (00.50)  \\
                       & GCN-MCD &  88.18 (00.47) &    10.00 &   5.56 (00.48) &   2.80 (00.26) &   3.51 (00.55) &   2.65 (00.47) &   3.43 (00.58)  \\
                       & GAT     &  91.03 (00.48) &    10.00 &   2.15 (00.41) &   2.28 (00.31) &   3.73 (00.45) &   2.17 (00.32) &   3.10 (00.33)  \\
                       & GAT-MCD &  90.84 (00.43) &    10.00 &   2.03 (00.40) &   2.21 (00.38) &   3.33 (00.45) &   2.58 (00.43) &   3.15 (00.44)  \\
                       & GIN     &  88.66 (00.45) &    10.00 &   3.69 (00.58) &   1.91 (00.35) &   4.26 (00.46) &   3.67 (00.50) &   4.04 (00.41)  \\
                       & GIN-MCD &  88.66 (00.42) &    10.00 &   7.86 (00.43) &   2.01 (00.47) &   4.06 (00.37) &   2.61 (00.29) &   3.16 (00.53)  \\
\midrule
\textsc{Amazon-Ph.}    & GCN     &  93.06 (00.36) &    12.50 &   1.50 (00.34) &   1.95 (00.56) &   2.29 (00.47) &   1.92 (00.49) &   1.75 (00.47)  \\
                       & GCN-MCD &  93.47 (00.37) &    12.50 &   1.74 (00.28) &   1.59 (00.29) &   2.41 (00.43) &   1.73 (00.23) &   1.95 (00.61)  \\
                       & GAT     &  93.41 (00.60) &    12.50 &   2.84 (00.58) &   1.71 (00.39) &   2.34 (00.42) &   1.67 (00.42) &   1.70 (00.22)  \\
                       & GAT-MCD &  93.11 (00.49) &    12.50 &   2.44 (00.42) &   1.90 (00.31) &   2.55 (00.49) &   1.89 (00.25) &   2.69 (00.54)  \\
                       & GIN     &  92.51 (00.63) &    12.50 &   2.89 (00.83) &   1.86 (00.56) &   2.34 (00.34) &   2.51 (00.45) &   3.24 (00.42)  \\
                       & GIN-MCD &  92.68 (00.53) &    12.50 &   2.32 (00.33) &   2.39 (00.33) &   2.35 (00.49) &   2.27 (00.41) &   2.69 (00.43)  \\
\bottomrule
\end{tabular}
}
\caption{
    \textbf{(\(\text{ECE}\), Accuracy, Not Balanced)}
    Accuracy and \(\text{ECE}\) results for all models in all dataset, as well as \(\text{ECE}\) results for calibration methods.
    For reference, accuracy of random model (1 / number of classes) is also provided.
    For each row, the best model (over all tested hyperparameter configurations), 
    was chosen using validation accuracy. Metrics shown in the table are not balanced.
}
\label{tab:results-ece-accuracy-not-balanced}
\end{table*}

\begin{table*}
\centering
\resizebox{\textwidth}{!}{
\begin{tabular}{llrrrrrrr}

\toprule
\multicolumn{9}{c}{Early Stopping using: Validation Accuracy, Test Metrics: Balanced} \\
\midrule
        &       & \multicolumn{2}{c}{ Accuracy (\%) }     &                                 \multicolumn{5}{c}{\(\text{ECE}\) (\%)} \\
\cmidrule(lr){3-4} \cmidrule(lr){5-9}
Dataset & Model & \multicolumn{1}{c}{Original} & \multicolumn{1}{c}{Random} & \multicolumn{1}{c}{Original}  & \multicolumn{1}{c}{Hist. Bin.}  & \multicolumn{1}{c}{Isotonic} & \multicolumn{1}{c}{TS w/ NLL} & \multicolumn{1}{c}{TS w/ Brier}  \\
\midrule
\textsc{Friendster}    & GCN     &  35.08 (02.32) &    25.00 &   4.08 (01.13) &  39.48 (00.24) &   5.70 (01.66) &   4.32 (01.27) &   4.87 (00.90)  \\
                       & GCN-MCD &  35.09 (00.94) &    25.00 &   4.40 (01.21) &  39.57 (00.17) &   6.43 (01.59) &   4.29 (01.37) &   5.20 (01.42)  \\
                       & GAT     &  32.40 (01.55) &    25.00 &   6.51 (01.88) &  40.54 (00.16) &   4.13 (00.89) &   3.63 (01.69) &   4.28 (01.19)  \\
                       & GAT-MCD &  31.24 (01.63) &    25.00 &   5.60 (01.64) &  40.66 (00.14) &   3.99 (00.66) &   3.56 (01.43) &   4.32 (01.13)  \\
                       & GIN     &  26.74 (00.84) &    25.00 &   3.43 (00.79) &  40.72 (00.08) &   7.44 (00.76) &   3.66 (00.64) &   4.50 (01.15)  \\
                       & GIN-MCD &  27.63 (01.13) &    25.00 &   3.18 (00.87) &  40.69 (00.16) &   7.00 (01.75) &   2.71 (00.81) &   2.88 (01.05)  \\
\midrule
\textsc{Cora}          & GCN     &  86.00 (01.25) &    14.29 &   5.31 (01.08) &   6.51 (00.86) &   6.66 (01.27) &   5.53 (01.05) &   5.57 (00.95)  \\
                       & GCN-MCD &  86.00 (01.02) &    14.29 &  10.23 (00.80) &   5.22 (01.20) &   4.88 (00.67) &   4.69 (00.79) &   5.14 (01.08)  \\
                       & GAT     &  86.31 (01.07) &    14.29 &  48.01 (01.00) &   5.72 (00.78) &   3.55 (00.62) &   4.56 (00.88) &   4.90 (00.65)  \\
                       & GAT-MCD &  85.30 (01.07) &    14.29 &  46.96 (01.22) &   4.35 (00.79) &   3.58 (00.70) &   5.01 (00.76) &   4.79 (00.92)  \\
                       & GIN     &  84.61 (00.80) &    14.29 &   8.76 (00.72) &   5.12 (01.10) &   5.71 (00.67) &   6.03 (01.20) &   6.90 (00.94)  \\
                       & GIN-MCD &  83.44 (01.49) &    14.29 &   5.18 (00.77) &   7.73 (01.16) &   5.54 (01.53) &   5.28 (01.21) &   5.51 (00.90)  \\
\midrule
\textsc{PubMed}        & GCN     &  87.63 (00.42) &    33.33 &   1.77 (00.30) &   1.85 (00.27) &   1.51 (00.34) &   1.25 (00.21) &   1.25 (00.21)  \\
                       & GCN-MCD &  87.67 (00.50) &    33.33 &   4.12 (00.45) &   2.06 (00.47) &   1.43 (00.38) &   1.35 (00.20) &   1.12 (00.24)  \\
                       & GAT     &  86.92 (00.50) &    33.33 &   3.42 (00.49) &   2.06 (00.32) &   1.87 (00.26) &   1.62 (00.26) &   1.64 (00.47)  \\
                       & GAT-MCD &  87.03 (00.45) &    33.33 &   5.07 (00.36) &   2.01 (00.31) &   2.01 (00.27) &   1.64 (00.28) &   1.75 (00.40)  \\
                       & GIN     &  86.66 (00.42) &    33.33 &   4.66 (00.37) &   1.97 (00.48) &   1.74 (00.31) &   1.68 (00.36) &   1.83 (00.33)  \\
                       & GIN-MCD &  86.30 (00.59) &    33.33 &   3.85 (00.36) &   2.00 (00.35) &   1.83 (00.50) &   2.21 (00.26) &   2.21 (00.32)  \\
\midrule
\textsc{CiteSeer}      & GCN     &  74.39 (01.76) &    16.67 &  17.71 (01.79) &   4.48 (01.02) &   5.56 (01.11) &   5.30 (00.85) &   5.81 (00.93)  \\
                       & GCN-MCD &  74.90 (01.32) &    16.67 &  23.53 (01.29) &   6.55 (01.26) &   6.45 (01.49) &   5.84 (01.14) &   5.22 (01.06)  \\
                       & GAT     &  73.57 (01.62) &    16.67 &  29.85 (01.64) &   6.35 (01.85) &   7.45 (01.43) &   7.11 (01.14) &   6.82 (01.75)  \\
                       & GAT-MCD &  72.05 (01.82) &    16.67 &  29.37 (01.91) &   5.82 (01.45) &   6.06 (01.20) &   5.84 (01.38) &   6.83 (00.67)  \\
                       & GIN     &  62.85 (01.70) &    16.67 &  11.75 (01.57) &   5.41 (00.94) &   6.05 (01.05) &   6.90 (01.24) &   6.49 (01.48)  \\
                       & GIN-MCD &  62.88 (01.24) &    16.67 &   5.67 (01.51) &   6.24 (01.14) &   6.92 (01.31) &   6.28 (00.56) &   5.95 (01.19)  \\
\midrule
\textsc{Cora-Full}     & GCN     &  69.62 (00.87) &     1.43 &   6.73 (00.98) &   7.82 (01.06) &   6.96 (00.52) &   5.18 (00.56) &   5.37 (00.94)  \\
                       & GCN-MCD &  69.49 (00.91) &     1.43 &   6.07 (00.49) &   7.61 (00.63) &   5.24 (00.66) &   4.15 (00.83) &   3.46 (00.71)  \\
                       & GAT     &  68.81 (01.17) &     1.43 &   4.62 (00.58) &   7.55 (00.70) &   5.88 (00.73) &   4.34 (00.73) &   4.34 (00.65)  \\
                       & GAT-MCD &  69.39 (00.91) &     1.43 &   3.95 (00.50) &   6.49 (00.82) &   5.52 (00.68) &   4.17 (00.48) &   3.61 (00.53)  \\
                       & GIN     &  66.71 (00.63) &     1.43 &  11.01 (01.01) &   6.90 (00.91) &   6.02 (00.37) &   9.69 (00.86) &   9.48 (00.53)  \\
                       & GIN-MCD &  65.84 (01.14) &     1.43 &  27.62 (00.92) &   5.80 (00.46) &   3.07 (00.61) &   6.25 (00.81) &   6.28 (00.69)  \\
\midrule
\textsc{Facebook}      & GCN     &  70.15 (01.44) &    50.00 &   6.16 (01.55) &  12.59 (01.03) &   7.55 (00.86) &   5.49 (00.87) &   5.64 (01.13)  \\
                       & GCN-MCD &  69.49 (01.08) &    50.00 &   5.31 (00.80) &  12.95 (01.18) &   5.73 (01.42) &   4.55 (01.08) &   5.50 (01.01)  \\
                       & GAT     &  67.19 (02.08) &    50.00 &   5.92 (01.57) &  15.99 (00.91) &   6.78 (01.09) &   4.02 (00.76) &   4.82 (01.07)  \\
                       & GAT-MCD &  68.62 (01.23) &    50.00 &   6.28 (00.93) &   8.92 (01.13) &   6.39 (01.04) &   5.75 (01.26) &   5.35 (01.15)  \\
                       & GIN     &  65.11 (01.24) &    50.00 &   5.61 (00.79) &  11.85 (02.08) &   6.14 (01.26) &   5.65 (01.48) &   5.64 (01.83)  \\
                       & GIN-MCD &  65.64 (01.68) &    50.00 &   4.78 (00.98) &  11.88 (01.57) &   6.37 (00.81) &   5.48 (01.58) &   6.21 (01.70)  \\
\midrule
\textsc{Amazon-Cmp.}   & GCN     &  91.51 (00.54) &    10.00 &   3.77 (00.36) &   2.83 (00.36) &   2.74 (00.41) &   1.98 (00.48) &   2.17 (00.40)  \\
                       & GCN-MCD &  91.51 (00.60) &    10.00 &   3.76 (00.59) &   2.07 (00.54) &   2.33 (00.60) &   1.81 (00.45) &   2.25 (00.29)  \\
                       & GAT     &  93.15 (00.46) &    10.00 &   1.80 (00.33) &   2.45 (00.64) &   2.43 (00.28) &   1.94 (00.44) &   2.73 (00.47)  \\
                       & GAT-MCD &  93.20 (00.26) &    10.00 &   2.26 (00.41) &   1.99 (00.34) &   2.25 (00.36) &   1.68 (00.35) &   2.79 (00.43)  \\
                       & GIN     &  92.11 (00.53) &    10.00 &   2.85 (00.37) &   4.18 (00.52) &   2.47 (00.34) &   2.35 (00.44) &   2.71 (00.50)  \\
                       & GIN-MCD &  92.08 (00.53) &    10.00 &   7.28 (00.46) &   3.52 (00.53) &   2.68 (00.39) &   2.27 (00.61) &   2.38 (00.35)  \\
\midrule
\textsc{Amazon-Ph.}    & GCN     &  93.61 (00.69) &    12.50 &   1.71 (00.32) &   2.37 (00.27) &   2.33 (00.35) &   2.23 (00.44) &   1.70 (00.46)  \\
                       & GCN-MCD &  93.68 (00.86) &    12.50 &   1.74 (00.34) &   2.17 (00.59) &   2.27 (00.51) &   2.19 (00.37) &   1.69 (00.31)  \\
                       & GAT     &  93.38 (00.77) &    12.50 &   3.13 (00.53) &   1.79 (00.32) &   2.09 (00.31) &   2.20 (00.51) &   1.84 (00.39)  \\
                       & GAT-MCD &  93.04 (00.47) &    12.50 &   2.54 (00.33) &   2.28 (00.44) &   2.28 (00.55) &   1.90 (00.39) &   2.10 (00.39)  \\
                       & GIN     &  93.96 (00.59) &    12.50 &   2.74 (00.36) &   2.33 (00.35) &   2.41 (00.42) &   2.72 (00.63) &   2.86 (00.57)  \\
                       & GIN-MCD &  93.87 (00.61) &    12.50 &   2.49 (00.55) &   2.73 (00.28) &   2.24 (00.50) &   2.74 (00.56) &   2.94 (00.67)  \\
\bottomrule
\end{tabular}
}
\caption{
    \textbf{(\(\text{ECE}\), Accuracy, Balanced)}
    Accuracy and \(\text{ECE}\) results for all models in all dataset, as well as \(\text{ECE}\) results for calibration methods.
    For reference, accuracy of random model (1 / number of classes) is also provided.
    For each row, the best model (over all tested hyperparameter configurations), 
    was chosen using validation accuracy. Metrics shown in the table are balanced.
}
\label{tab:results-ece-accuracy-balanced}
\end{table*}

\begin{table*}
\centering
\resizebox{\textwidth}{!}{
\begin{tabular}{llrrrrrrr}

\toprule
\multicolumn{9}{c}{Early Stopping using: Validation Loss, Test Metrics: Not Balanced} \\
\midrule
        &       & \multicolumn{2}{c}{ Accuracy (\%) }     &                                 \multicolumn{5}{c}{\(\text{ECE}\) (\%)} \\
\cmidrule(lr){3-4} \cmidrule(lr){5-9}
Dataset & Model & \multicolumn{1}{c}{Original} & \multicolumn{1}{c}{Random} & \multicolumn{1}{c}{Original}  & \multicolumn{1}{c}{Hist. Bin.}  & \multicolumn{1}{c}{Isotonic} & \multicolumn{1}{c}{TS w/ NLL} & \multicolumn{1}{c}{TS w/ Brier}  \\
\midrule
\textsc{Friendster}    & GCN     &  31.97 (00.60) &    25.00 &   6.26 (00.72) &   3.27 (00.51) &  12.10 (00.43) &   7.06 (00.33) &   7.26 (00.40) \\
                       & GCN-MCD &  31.17 (00.34) &    25.00 &   4.95 (00.44) &   3.13 (00.64) &  15.22 (00.52) &   6.77 (00.44) &   6.38 (00.43) \\
                       & GAT     &  24.67 (00.61) &    25.00 &   7.39 (00.66) &   2.03 (00.31) &  17.94 (00.61) &   5.80 (00.37) &   6.03 (00.74) \\
                       & GAT-MCD &  27.83 (00.51) &    25.00 &   4.42 (00.42) &   1.77 (00.36) &  10.87 (00.49) &   3.81 (00.30) &   3.91 (00.46) \\
                       & GIN     &  23.75 (00.39) &    25.00 &   8.00 (00.38) &   1.39 (00.47) &  11.06 (00.52) &   7.99 (00.53) &   8.26 (00.51) \\
                       & GIN-MCD &  24.23 (00.58) &    25.00 &   7.87 (00.66) &   1.31 (00.70) &  15.57 (00.57) &   8.06 (00.31) &   7.71 (00.58) \\
\midrule
\textsc{Cora}          & GCN     &  86.19 (01.12) &    14.29 &   5.19 (00.75) &   5.87 (01.17) &   6.67 (01.08) &   5.08 (00.87) &   5.72 (00.83) \\
                       & GCN-MCD &  85.30 (00.90) &    14.29 &  10.19 (01.03) &   4.83 (00.78) &   5.58 (00.57) &   4.03 (00.75) &   4.49 (00.95) \\
                       & GAT     &  85.46 (01.14) &    14.29 &   4.26 (00.77) &   4.98 (01.07) &   5.66 (00.78) &   4.99 (00.58) &   3.98 (01.02) \\
                       & GAT-MCD &  84.81 (01.01) &    14.29 &   3.92 (00.47) &   6.15 (01.43) &   6.00 (01.23) &   4.87 (00.89) &   4.18 (00.93) \\
                       & GIN     &  83.64 (00.72) &    14.29 &   6.11 (01.01) &   6.03 (00.89) &   5.61 (00.97) &   6.12 (01.25) &   5.57 (00.60) \\
                       & GIN-MCD &  84.14 (01.63) &    14.29 &   6.58 (00.72) &   5.25 (01.04) &   5.50 (00.96) &   6.11 (00.60) &   6.46 (01.19) \\
\midrule
\textsc{PubMed}        & GCN     &  87.96 (00.45) &    33.33 &   1.85 (00.39) &   1.58 (00.40) &   1.35 (00.22) &   1.72 (00.28) &   1.76 (00.31) \\
                       & GCN-MCD &  87.98 (00.35) &    33.33 &   3.38 (00.43) &   1.50 (00.20) &   1.63 (00.47) &   1.43 (00.20) &   1.58 (00.35) \\
                       & GAT     &  87.71 (00.49) &    33.33 &   2.02 (00.39) &   2.30 (00.41) &   1.73 (00.37) &   1.69 (00.56) &   1.37 (00.38) \\
                       & GAT-MCD &  87.93 (00.57) &    33.33 &   1.53 (00.29) &   1.94 (00.32) &   1.95 (00.21) &   1.65 (00.36) &   1.44 (00.29) \\
                       & GIN     &  87.31 (00.69) &    33.33 &   1.36 (00.29) &   1.55 (00.28) &   1.76 (00.42) &   1.35 (00.40) &   1.19 (00.17) \\
                       & GIN-MCD &  87.23 (00.43) &    33.33 &   1.39 (00.20) &   1.65 (00.25) &   1.71 (00.36) &   1.26 (00.26) &   1.61 (00.27) \\
\midrule
\textsc{CiteSeer}      & GCN     &  76.92 (01.15) &    16.67 &   5.10 (01.20) &   5.12 (00.84) &   5.41 (01.12) &   4.37 (01.06) &   5.03 (01.11) \\
                       & GCN-MCD &  77.57 (01.06) &    16.67 &  12.25 (00.99) &   4.80 (01.01) &   5.57 (00.69) &   4.41 (01.07) &   5.02 (00.88) \\
                       & GAT     &  75.50 (01.46) &    16.67 &   9.60 (01.21) &   4.80 (00.92) &   6.12 (01.24) &   4.64 (00.80) &   4.41 (00.72) \\
                       & GAT-MCD &  74.66 (01.67) &    16.67 &   9.97 (01.15) &   4.25 (01.30) &   5.98 (00.93) &   3.82 (00.79) &   4.20 (00.70) \\
                       & GIN     &  69.24 (01.42) &    16.67 &   6.89 (01.35) &   6.30 (01.03) &   6.76 (01.23) &   6.87 (01.23) &   6.36 (00.84) \\
                       & GIN-MCD &  68.86 (01.47) &    16.67 &   6.76 (01.48) &   5.64 (01.11) &   6.02 (01.04) &   6.56 (01.08) &   6.85 (00.78) \\
\midrule
\textsc{Cora-Full}     & GCN     &  69.87 (00.61) &     1.43 &   4.45 (00.29) &   5.47 (00.67) &   4.91 (00.63) &   4.68 (00.50) &   4.31 (00.27) \\
                       & GCN-MCD &  69.93 (00.74) &     1.43 &  14.91 (00.57) &   2.93 (00.32) &   2.88 (00.41) &   3.48 (00.44) &   3.28 (00.52) \\
                       & GAT     &  69.67 (00.46) &     1.43 &   3.78 (00.28) &   4.56 (00.48) &   4.21 (00.38) &   4.43 (00.43) &   4.02 (00.35) \\
                       & GAT-MCD &  69.14 (00.72) &     1.43 &   3.04 (00.72) &   4.49 (00.53) &   4.14 (00.50) &   3.02 (00.53) &   3.17 (00.59) \\
                       & GIN     &  63.50 (00.85) &     1.43 &   4.62 (00.79) &   5.29 (00.50) &   3.84 (00.53) &   7.28 (00.42) &   3.83 (00.59) \\
                       & GIN-MCD &  63.30 (00.60) &     1.43 &  19.35 (00.65) &   4.81 (00.73) &   3.70 (00.58) &   3.99 (00.59) &   2.59 (00.56) \\
\midrule
\textsc{Facebook}      & GCN     &  68.87 (01.38) &    50.00 &   5.88 (01.18) &   6.31 (01.47) &   9.20 (01.26) &   6.01 (00.95) &   4.94 (01.03) \\
                       & GCN-MCD &  69.30 (01.27) &    50.00 &   6.37 (01.15) &   5.51 (01.04) &   8.16 (00.75) &   5.99 (00.83) &   5.86 (00.80) \\
                       & GAT     &  65.02 (01.83) &    50.00 &   4.43 (01.03) &   4.55 (01.04) &   7.68 (01.38) &   4.29 (01.17) &   4.19 (01.04) \\
                       & GAT-MCD &  67.07 (01.81) &    50.00 &   5.39 (01.15) &   2.98 (01.02) &   7.56 (01.36) &   4.57 (01.02) &   3.79 (00.83) \\
                       & GIN     &  66.33 (01.93) &    50.00 &   4.83 (01.87) &   3.98 (00.97) &   8.85 (00.90) &   4.40 (00.91) &   4.51 (01.11) \\
                       & GIN-MCD &  66.06 (01.91) &    50.00 &   3.77 (00.95) &   3.40 (00.96) &   9.67 (01.27) &   4.60 (00.80) &   5.01 (00.97) \\
\midrule
\textsc{Amazon-Cmp.}   & GCN     &  87.59 (00.65) &    10.00 &   4.33 (00.59) &   1.81 (00.36) &   4.05 (00.47) &   2.80 (00.39) &   3.33 (00.41) \\
                       & GCN-MCD &  87.80 (00.30) &    10.00 &   4.62 (00.43) &   2.08 (00.34) &   3.90 (00.65) &   2.38 (00.50) &   3.38 (00.48) \\
                       & GAT     &  89.42 (00.52) &    10.00 &   1.95 (00.50) &   2.09 (00.39) &   4.07 (00.40) &   2.34 (00.43) &   2.18 (00.43) \\
                       & GAT-MCD &  89.63 (00.64) &    10.00 &   1.63 (00.52) &   2.19 (00.37) &   3.51 (00.69) &   2.18 (00.57) &   1.87 (00.30) \\
                       & GIN     &  89.28 (00.68) &    10.00 &   3.72 (00.54) &   1.74 (00.41) &   4.44 (00.43) &   3.59 (00.27) &   3.60 (00.44) \\
                       & GIN-MCD &  89.17 (00.65) &    10.00 &   3.65 (00.46) &   1.90 (00.26) &   4.49 (00.41) &   3.54 (00.44) &   3.46 (00.19) \\
\midrule
\textsc{Amazon-Ph.}    & GCN     &  93.36 (00.61) &    12.50 &   2.21 (00.41) &   2.03 (00.29) &   2.18 (00.51) &   2.23 (00.46) &   2.25 (00.53) \\
                       & GCN-MCD &  93.01 (00.69) &    12.50 &   2.13 (00.31) &   1.74 (00.35) &   1.81 (00.29) &   2.21 (00.52) &   2.17 (00.57) \\
                       & GAT     &  93.27 (00.57) &    12.50 &   1.67 (00.23) &   2.01 (00.63) &   1.76 (00.39) &   1.71 (00.31) &   1.90 (00.49) \\
                       & GAT-MCD &  93.38 (00.46) &    12.50 &   2.07 (00.37) &   2.13 (00.48) &   1.72 (00.25) &   1.78 (00.32) &   1.88 (00.26) \\
                       & GIN     &  92.53 (00.75) &    12.50 &   3.36 (00.72) &   2.69 (00.65) &   2.92 (00.51) &   2.40 (00.30) &   3.68 (00.49) \\
                       & GIN-MCD &  92.20 (00.52) &    12.50 &   1.97 (00.26) &   2.17 (00.57) &   2.78 (00.47) &   2.25 (00.41) &   2.82 (00.43) \\
\bottomrule
\end{tabular}
}
\caption{
    \textbf{(\(\text{ECE}\), Loss, Not Balanced)}
    Accuracy and \(\text{ECE}\) results for all models in all dataset, as well as \(\text{ECE}\) results for calibration methods.
    For reference, accuracy of random model (1 / number of classes) is also provided.
    For each row, the best model (over all tested hyperparameter configurations), 
    was chosen using validation loss. Metrics shown in the table are not balanced.
}
\label{tab:results-ece-loss-not-balanced}
\end{table*}

\begin{table*}
\centering
\resizebox{\textwidth}{!}{
\begin{tabular}{llrrrrrrr}

\toprule
\multicolumn{9}{c}{Early Stopping using: Validation Loss, Test Metrics: Balanced} \\
\midrule
        &       & \multicolumn{2}{c}{ Accuracy (\%) }     &                                 \multicolumn{5}{c}{\(\text{ECE}\) (\%)} \\
\cmidrule(lr){3-4} \cmidrule(lr){5-9}
Dataset & Model & \multicolumn{1}{c}{Original} & \multicolumn{1}{c}{Random} & \multicolumn{1}{c}{Original}  & \multicolumn{1}{c}{Hist. Bin.}  & \multicolumn{1}{c}{Isotonic} & \multicolumn{1}{c}{TS w/ NLL} & \multicolumn{1}{c}{TS w/ Brier} \\
\midrule
\textsc{Friendster}    & GCN     &  35.11 (01.35) &    25.00 &   7.03 (01.61) &  39.81 (00.24) &   4.78 (01.50) &   5.11 (01.35) &   5.88 (01.37) \\
                       & GCN-MCD &  33.31 (01.21) &    25.00 &   4.16 (01.01) &  39.60 (00.14) &   5.47 (01.93) &   3.53 (01.33) &   3.93 (01.11) \\
                       & GAT     &  30.71 (01.87) &    25.00 &   4.19 (01.36) &  40.29 (00.18) &   4.70 (00.82) &   4.36 (01.33) &   4.23 (01.45) \\
                       & GAT-MCD &  28.81 (01.44) &    25.00 &   4.20 (01.43) &  40.14 (00.16) &   5.93 (00.88) &   3.44 (01.46) &   3.67 (01.15) \\
                       & GIN     &  26.82 (00.91) &    25.00 &   3.17 (00.87) &  40.78 (00.14) &   7.01 (01.49) &   3.07 (00.50) &   3.83 (00.90) \\
                       & GIN-MCD &  26.90 (00.82) &    25.00 &   3.48 (00.74) &  40.60 (00.13) &   7.21 (01.40) &   3.30 (00.69) &   4.08 (00.71) \\
\midrule
\textsc{Cora}          & GCN     &  86.73 (01.81) &    14.29 &   4.32 (00.84) &   7.06 (01.08) &   6.84 (01.05) &   5.42 (01.11) &   5.91 (01.05) \\
                       & GCN-MCD &  85.50 (01.10) &    14.29 &   9.91 (01.34) &   6.75 (01.21) &   6.27 (00.56) &   3.85 (00.91) &   5.07 (00.96) \\
                       & GAT     &  84.24 (00.99) &    14.29 &   4.95 (00.55) &   6.16 (00.87) &   6.33 (00.75) &   5.60 (00.64) &   5.48 (00.91) \\
                       & GAT-MCD &  85.10 (01.67) &    14.29 &   4.06 (00.72) &   6.80 (00.85) &   5.36 (00.80) &   4.79 (01.03) &   4.54 (01.00) \\
                       & GIN     &  83.76 (01.19) &    14.29 &   6.80 (01.48) &   6.78 (00.50) &   5.51 (01.03) &   6.07 (01.13) &   6.21 (01.24) \\
                       & GIN-MCD &  83.00 (01.00) &    14.29 &   6.32 (00.96) &   4.71 (00.99) &   4.83 (00.65) &   6.26 (01.04) &   6.71 (01.21) \\
\midrule
\textsc{PubMed}        & GCN     &  88.73 (00.57) &    33.33 &   1.83 (00.28) &   1.60 (00.35) &   1.53 (00.35) &   1.80 (00.35) &   1.65 (00.40) \\
                       & GCN-MCD &  88.47 (00.44) &    33.33 &   3.77 (00.22) &   1.64 (00.27) &   1.48 (00.27) &   1.51 (00.30) &   1.53 (00.26) \\
                       & GAT     &  88.19 (00.36) &    33.33 &   1.74 (00.23) &   2.22 (00.32) &   1.35 (00.30) &   1.53 (00.22) &   1.39 (00.36) \\
                       & GAT-MCD &  87.72 (00.38) &    33.33 &   1.49 (00.20) &   2.11 (00.38) &   1.31 (00.30) &   1.53 (00.29) &   1.27 (00.16) \\
                       & GIN     &  87.38 (00.54) &    33.33 &   1.42 (00.37) &   2.00 (00.35) &   1.60 (00.39) &   1.25 (00.27) &   1.39 (00.32) \\
                       & GIN-MCD &  87.31 (00.60) &    33.33 &   1.38 (00.44) &   2.44 (00.23) &   1.54 (00.26) &   1.37 (00.42) &   1.44 (00.31) \\
\midrule
\textsc{CiteSeer}      & GCN     &  74.53 (02.07) &    16.67 &   5.57 (01.16) &   6.49 (01.63) &   6.02 (00.96) &   4.89 (01.07) &   6.12 (01.05) \\
                       & GCN-MCD &  75.03 (02.35) &    16.67 &  11.33 (02.26) &   7.48 (01.37) &   6.59 (01.75) &   5.56 (01.16) &   5.60 (01.40) \\
                       & GAT     &  74.72 (01.72) &    16.67 &   9.63 (01.73) &   6.77 (01.13) &   6.64 (00.99) &   5.65 (00.85) &   5.12 (00.77) \\
                       & GAT-MCD &  74.16 (01.64) &    16.67 &   9.60 (01.57) &   6.16 (00.84) &   6.34 (01.03) &   5.36 (01.00) &   5.55 (00.70) \\
                       & GIN     &  69.49 (01.33) &    16.67 &   7.29 (01.65) &   5.49 (01.22) &   7.48 (01.37) &   7.07 (01.65) &   7.70 (00.87) \\
                       & GIN-MCD &  68.66 (01.33) &    16.67 &   7.88 (01.07) &   5.57 (01.18) &   7.54 (01.17) &   7.15 (00.98) &   6.88 (01.09) \\
\midrule
\textsc{Cora-Full}     & GCN     &  68.98 (01.05) &     1.43 &   4.25 (00.90) &   7.77 (00.72) &   6.38 (00.99) &   4.42 (00.54) &   4.21 (00.40) \\
                       & GCN-MCD &  68.98 (00.90) &     1.43 &  13.43 (00.92) &   6.53 (00.93) &   3.32 (00.61) &   4.58 (00.54) &   4.08 (00.76) \\
                       & GAT     &  68.85 (00.93) &     1.43 &   4.66 (00.47) &   7.83 (01.05) &   5.38 (00.70) &   4.01 (00.37) &   4.23 (00.58) \\
                       & GAT-MCD &  68.89 (01.24) &     1.43 &   3.80 (00.91) &   7.12 (00.74) &   5.25 (00.91) &   3.48 (00.91) &   4.08 (00.42) \\
                       & GIN     &  62.45 (01.13) &     1.43 &   3.55 (00.87) &   6.77 (01.02) &   5.00 (00.41) &   5.27 (00.70) &   4.61 (00.54) \\
                       & GIN-MCD &  62.76 (00.84) &     1.43 &  17.81 (00.66) &   5.18 (00.88) &   4.80 (00.75) &   3.68 (00.54) &   3.51 (00.42) \\
\midrule
\textsc{Facebook}      & GCN     &  70.08 (01.37) &    50.00 &   5.45 (01.11) &   9.79 (00.56) &   7.30 (00.94) &   5.06 (00.61) &   5.84 (01.02) \\
                       & GCN-MCD &  69.62 (01.80) &    50.00 &   5.62 (01.17) &   8.77 (01.14) &   6.99 (01.20) &   6.00 (01.33) &   5.87 (01.07) \\
                       & GAT     &  66.15 (01.42) &    50.00 &   3.38 (00.85) &   8.45 (00.97) &   4.39 (01.18) &   3.68 (00.94) &   3.87 (01.43) \\
                       & GAT-MCD &  66.57 (01.62) &    50.00 &   4.37 (01.64) &   9.88 (01.11) &   4.12 (01.22) &   4.35 (01.13) &   4.48 (00.93) \\
                       & GIN     &  64.93 (01.45) &    50.00 &   5.52 (01.17) &  11.09 (01.10) &   6.31 (00.84) &   4.92 (00.62) &   5.29 (01.04) \\
                       & GIN-MCD &  64.78 (01.34) &    50.00 &   4.46 (00.84) &  12.23 (00.73) &   6.01 (01.14) &   5.19 (01.28) &   5.03 (01.21) \\
\midrule
\textsc{Amazon-Cmp.}   & GCN     &  91.31 (00.64) &    10.00 &   3.38 (00.21) &   1.92 (00.60) &   2.51 (00.43) &   2.05 (00.29) &   2.57 (00.35) \\
                       & GCN-MCD &  91.60 (00.49) &    10.00 &   3.83 (00.36) &   1.98 (00.58) &   2.29 (00.27) &   1.69 (00.30) &   2.44 (00.42) \\
                       & GAT     &  92.87 (00.60) &    10.00 &   1.78 (00.42) &   2.21 (00.57) &   2.68 (00.43) &   2.32 (00.49) &   2.17 (00.59) \\
                       & GAT-MCD &  92.54 (00.65) &    10.00 &   1.59 (00.35) &   2.52 (00.42) &   2.65 (00.48) &   1.76 (00.24) &   1.96 (00.26) \\
                       & GIN     &  91.10 (00.48) &    10.00 &   2.84 (00.43) &   3.08 (00.66) &   2.82 (00.60) &   2.63 (00.43) &   2.48 (00.40) \\
                       & GIN-MCD &  91.33 (00.71) &    10.00 &   2.63 (00.43) &   2.34 (00.35) &   2.67 (00.40) &   2.63 (00.59) &   2.86 (00.44) \\
\midrule
\textsc{Amazon-Ph.}    & GCN     &  94.24 (00.58) &    12.50 &   2.20 (00.32) &   2.44 (00.58) &   1.79 (00.30) &   2.34 (00.44) &   1.90 (00.41) \\
                       & GCN-MCD &  94.12 (00.58) &    12.50 &   2.45 (00.26) &   2.53 (00.49) &   2.05 (00.43) &   2.36 (00.43) &   1.85 (00.40) \\
                       & GAT     &  93.91 (00.47) &    12.50 &   1.76 (00.42) &   2.92 (00.59) &   1.78 (00.39) &   2.06 (00.32) &   2.17 (00.64) \\
                       & GAT-MCD &  93.33 (00.71) &    12.50 &   2.44 (00.57) &   2.82 (00.67) &   1.83 (00.63) &   2.19 (00.46) &   2.20 (00.37) \\
                       & GIN     &  93.31 (00.79) &    12.50 &   3.06 (00.58) &   3.36 (00.71) &   3.10 (00.57) &   2.30 (00.51) &   3.19 (00.17) \\
                       & GIN-MCD &  93.04 (00.40) &    12.50 &   2.27 (00.48) &   3.05 (00.87) &   2.68 (00.80) &   2.51 (00.29) &   2.87 (00.46) \\
\bottomrule
\end{tabular}
}
\caption{
    \textbf{(\(\text{ECE}\), Loss, Balanced)}
    Accuracy and \(\text{ECE}\) results for all models in all dataset, as well as \(\text{ECE}\) results for calibration methods.
    For reference, accuracy of random model (1 / number of classes) is also provided.
    For each row, the best model (over all tested hyperparameter configurations), 
    was chosen using validation loss. Metrics shown in the table are balanced.
}
\label{tab:results-ece-loss-balanced}
\end{table*}


\begin{table*}
\centering
\resizebox{\textwidth}{!}{
\begin{tabular}{llrrrrrrrr}

\toprule
\multicolumn{9}{c}{Early Stopping using: Validation Accuracy, Test Metrics: Not Balanced} \\
\midrule
        &       & \multicolumn{2}{c}{ Accuracy (\%) }     &                                 \multicolumn{5}{c}{\(\text{ECE}_{\geq 50}\) (\%)} \\
\cmidrule(lr){3-4} \cmidrule(lr){5-9}
Dataset & Model & \multicolumn{1}{c}{Original} & \multicolumn{1}{c}{Random} & \multicolumn{1}{c}{Original}  & \multicolumn{1}{c}{Hist. Bin.}  & \multicolumn{1}{c}{Isotonic} & \multicolumn{1}{c}{TS w/ NLL} & \multicolumn{1}{c}{TS w/ Brier} \\
\midrule
\textsc{Friendster}    & GCN     &  39.70 (00.56) &    25.00 &            32.55 (01.98) &   3.71 (00.46) &  37.62 (02.51) &  31.04 (04.31) &  30.69 (02.06) \\
                       & GCN-MCD &  40.48 (00.69) &    25.00 &            26.36 (02.17) &   4.65 (00.51) &  28.71 (01.53) &  27.51 (02.62) &  28.71 (02.16) \\
                       & GAT     &  29.03 (00.40) &    25.00 &            48.89 (02.89) &   2.68 (00.47) &  43.16 (02.13) &  49.07 (02.51) &  52.88 (01.75) \\
                       & GAT-MCD &  28.53 (00.59) &    25.00 &            45.76 (03.20) &   3.29 (00.69) &  42.02 (02.77) &  43.25 (03.54) &  46.95 (04.30) \\
                       & GIN     &  23.84 (00.54) &    25.00 &            42.58 (07.37) &   1.27 (00.59) &  47.85 (01.89) &  50.24 (08.94) &  44.74 (06.77) \\
                       & GIN-MCD &  23.91 (00.45) &    25.00 &            46.11 (06.00) &   1.39 (00.45) &  47.62 (01.14) &  46.48 (06.67) &  48.77 (06.25) \\
\midrule
\textsc{Cora}          & GCN     &  86.03 (01.25) &    14.29 &             3.83 (01.14) &   5.74 (01.00) &   6.28 (00.74) &   4.10 (00.95) &   5.87 (00.73) \\
                       & GCN-MCD &  86.06 (00.56) &    14.29 &             9.82 (00.59) &   4.68 (00.49) &   5.15 (01.17) &   3.70 (00.71) &   4.46 (00.84) \\
                       & GAT     &  85.29 (00.88) &    14.29 &            40.25 (01.13) &   5.83 (01.04) &   3.15 (00.99) &   3.92 (00.73) &   3.76 (00.92) \\
                       & GAT-MCD &  84.79 (01.01) &    14.29 &            37.35 (01.01) &   3.80 (00.77) &   2.39 (00.68) &   4.33 (00.70) &   4.61 (00.69) \\
                       & GIN     &  83.35 (01.14) &    14.29 &             9.33 (01.22) &   5.79 (01.36) &   5.19 (00.97) &   4.71 (01.25) &   5.52 (01.15) \\
                       & GIN-MCD &  83.77 (00.85) &    14.29 &             3.35 (00.65) &   5.85 (00.94) &   4.88 (00.71) &   4.32 (01.33) &   3.96 (00.90) \\
\midrule
\textsc{PubMed}        & GCN     &  87.31 (00.57) &    33.33 &             1.45 (00.33) &   1.57 (00.30) &   1.84 (00.31) &   1.21 (00.37) &   1.10 (00.35) \\
                       & GCN-MCD &  87.51 (00.48) &    33.33 &             3.98 (00.35) &   2.03 (00.28) &   1.82 (00.27) &   1.35 (00.44) &   1.28 (00.47) \\
                       & GAT     &  86.71 (00.35) &    33.33 &             3.18 (00.37) &   1.54 (00.37) &   1.50 (00.25) &   1.42 (00.26) &   1.34 (00.32) \\
                       & GAT-MCD &  86.53 (00.25) &    33.33 &             4.84 (00.30) &   1.47 (00.26) &   1.75 (00.31) &   1.18 (00.31) &   1.32 (00.20) \\
                       & GIN     &  86.93 (00.40) &    33.33 &             4.59 (00.46) &   1.45 (00.45) &   1.74 (00.24) &   1.73 (00.36) &   1.63 (00.33) \\
                       & GIN-MCD &  86.84 (00.21) &    33.33 &             3.46 (00.25) &   1.30 (00.30) &   2.23 (00.36) &   1.76 (00.28) &   2.16 (00.26) \\
\midrule
\textsc{CiteSeer}      & GCN     &  77.21 (01.05) &    16.67 &            20.15 (01.13) &   4.56 (00.98) &   4.68 (01.11) &   4.43 (01.30) &   5.07 (00.92) \\
                       & GCN-MCD &  77.32 (01.48) &    16.67 &            25.80 (01.34) &   3.42 (00.70) &   4.17 (00.94) &   3.87 (00.97) &   4.30 (01.33) \\
                       & GAT     &  75.99 (01.11) &    16.67 &            30.14 (01.31) &   5.34 (01.13) &   6.90 (01.13) &   5.50 (00.87) &   3.68 (00.67) \\
                       & GAT-MCD &  76.49 (01.22) &    16.67 &            31.56 (01.52) &   5.59 (00.80) &   4.33 (00.61) &   6.25 (01.32) &   5.56 (01.32) \\
                       & GIN     &  67.00 (00.83) &    16.67 &             8.75 (00.85) &   5.10 (00.93) &   6.84 (01.09) &   6.08 (01.18) &   7.10 (01.59) \\
                       & GIN-MCD &  66.87 (01.47) &    16.67 &             7.60 (01.20) &   5.62 (01.49) &   6.73 (00.78) &   7.52 (01.50) &   7.39 (01.48) \\
\midrule
\textsc{Cora-Full}     & GCN     &  70.34 (00.61) &     1.43 &             6.49 (00.59) &   5.62 (00.57) &   6.05 (00.57) &   3.99 (00.59) &   4.44 (00.47) \\
                       & GCN-MCD &  70.31 (00.55) &     1.43 &             4.70 (00.36) &   3.83 (00.58) &   3.56 (00.52) &   3.01 (00.49) &   2.92 (00.63) \\
                       & GAT     &  69.44 (00.51) &     1.43 &             3.02 (00.59) &   4.48 (00.62) &   4.65 (00.40) &   3.16 (00.41) &   3.02 (00.65) \\
                       & GAT-MCD &  69.90 (00.54) &     1.43 &             2.36 (00.74) &   3.63 (00.54) &   4.11 (00.30) &   2.54 (00.59) &   2.59 (00.33) \\
                       & GIN     &  68.84 (00.45) &     1.43 &             6.52 (00.57) &   5.52 (00.67) &   5.48 (00.47) &   6.09 (00.27) &   6.34 (00.65) \\
                       & GIN-MCD &  68.94 (00.77) &     1.43 &            27.21 (00.68) &   6.85 (00.53) &   3.57 (00.45) &   3.31 (00.66) &   3.33 (00.50) \\
\midrule
\textsc{Facebook}      & GCN     &  69.07 (01.49) &    50.00 &             4.93 (00.84) &   5.21 (01.01) &   9.81 (01.31) &   4.88 (01.50) &   5.15 (01.08) \\
                       & GCN-MCD &  71.44 (00.91) &    50.00 &             6.98 (00.84) &   5.59 (00.70) &   8.67 (00.95) &   4.67 (01.41) &   5.20 (00.95) \\
                       & GAT     &  67.45 (01.64) &    50.00 &             5.58 (01.72) &   5.36 (00.91) &  17.37 (00.73) &   4.21 (01.10) &   4.75 (00.94) \\
                       & GAT-MCD &  68.47 (01.27) &    50.00 &             5.94 (00.41) &   6.31 (01.42) &  11.83 (01.35) &   6.23 (00.98) &   5.27 (01.12) \\
                       & GIN     &  66.11 (01.97) &    50.00 &             5.81 (01.87) &   3.73 (01.02) &   8.92 (01.02) &   4.45 (00.89) &   5.52 (01.26) \\
                       & GIN-MCD &  65.56 (01.58) &    50.00 &             4.18 (01.16) &   4.84 (00.85) &   8.10 (01.71) &   4.76 (00.98) &   4.98 (01.13) \\
\midrule
\textsc{Amazon-Cmp.}   & GCN     &  87.97 (00.57) &    10.00 &             4.25 (00.44) &   1.54 (00.39) &   3.55 (00.20) &   2.55 (00.46) &   3.17 (00.50) \\
                       & GCN-MCD &  88.18 (00.47) &    10.00 &             4.54 (00.40) &   2.50 (00.31) &   3.39 (00.62) &   2.35 (00.47) &   3.25 (00.55) \\
                       & GAT     &  91.03 (00.48) &    10.00 &             2.03 (00.42) &   2.27 (00.29) &   3.44 (00.46) &   2.04 (00.27) &   3.05 (00.29) \\
                       & GAT-MCD &  90.84 (00.43) &    10.00 &             1.79 (00.37) &   2.11 (00.37) &   3.23 (00.44) &   2.47 (00.42) &   3.04 (00.42) \\
                       & GIN     &  88.66 (00.45) &    10.00 &             2.53 (00.37) &   1.58 (00.40) &   3.99 (00.49) &   2.72 (00.40) &   3.71 (00.37) \\
                       & GIN-MCD &  88.66 (00.42) &    10.00 &             6.95 (00.41) &   1.83 (00.43) &   3.68 (00.40) &   2.20 (00.33) &   3.01 (00.54) \\
\midrule
\textsc{Amazon-Ph.}    & GCN     &  93.06 (00.36) &    12.50 &             1.23 (00.36) &   1.80 (00.52) &   2.07 (00.50) &   1.57 (00.45) &   1.47 (00.40) \\
                       & GCN-MCD &  93.47 (00.37) &    12.50 &             1.44 (00.25) &   1.45 (00.27) &   2.02 (00.49) &   1.25 (00.30) &   1.64 (00.54) \\
                       & GAT     &  93.41 (00.60) &    12.50 &             2.55 (00.57) &   1.47 (00.37) &   2.22 (00.37) &   1.32 (00.37) &   1.43 (00.21) \\
                       & GAT-MCD &  93.11 (00.49) &    12.50 &             2.27 (00.43) &   1.79 (00.30) &   2.22 (00.45) &   1.60 (00.26) &   2.39 (00.56) \\
                       & GIN     &  92.51 (00.63) &    12.50 &             2.21 (00.69) &   1.60 (00.54) &   1.94 (00.32) &   1.52 (00.44) &   2.81 (00.52) \\
                       & GIN-MCD &  92.68 (00.53) &    12.50 &             1.42 (00.35) &   2.18 (00.35) &   2.11 (00.46) &   1.35 (00.28) &   2.14 (00.44) \\
\bottomrule
\end{tabular}
}
\caption{
    \textbf{(\(\text{ECE}_{\geq 50}\), Accuracy, Not Balanced)}
    Accuracy and \(\text{ECE}_{\geq 50}\) results for all models in all dataset, as well as \(\text{ECE}_{\geq 50}\) results for calibration methods.
    For reference, accuracy of random model (1 / number of classes) is also provided.
    For each row, the best model (over all tested hyperparameter configurations), 
    was chosen using validation accuracy. Metrics shown in the table are not balanced.
}
\label{tab:results-ece-50-accuracy-not-balanced}
\end{table*}

\begin{table*}
\centering
\resizebox{\textwidth}{!}{
\begin{tabular}{llrrrrrrr}

\toprule
\multicolumn{9}{c}{Early Stopping using: Validation Accuracy, Test Metrics: Balanced} \\
\midrule
        &       & \multicolumn{2}{c}{ Accuracy (\%) }     &                                 \multicolumn{5}{c}{\(\text{ECE}_{\geq 50}\) (\%)} \\
\cmidrule(lr){3-4} \cmidrule(lr){5-9}
Dataset & Model & \multicolumn{1}{c}{Original} & \multicolumn{1}{c}{Random} & \multicolumn{1}{c}{Original}  & \multicolumn{1}{c}{Hist. Bin.}  & \multicolumn{1}{c}{Isotonic} & \multicolumn{1}{c}{TS w/ NLL} & \multicolumn{1}{c}{TS w/ Brier} \\
\midrule
\textsc{Friendster}    & GCN     &  35.08 (02.32) &    25.00 &            16.08 (02.80) &  39.73 (00.29) &  18.52 (05.94) &  23.61 (06.17) &  16.92 (03.80) \\
                       & GCN-MCD &  35.09 (00.94) &    25.00 &            17.18 (04.40) &  39.98 (00.18) &  16.46 (05.34) &  17.12 (05.02) &  16.49 (03.63) \\
                       & GAT     &  32.40 (01.55) &    25.00 &            35.48 (04.78) &  40.59 (00.16) &  39.13 (04.83) &  36.50 (07.71) &  33.85 (08.56) \\
                       & GAT-MCD &  31.24 (01.63) &    25.00 &            34.83 (06.91) &  40.76 (00.17) &  28.52 (07.38) &  29.64 (06.34) &  31.69 (08.26) \\
                       & GIN     &  26.74 (00.84) &    25.00 &            33.25 (07.94) &  40.72 (00.08) &  43.93 (01.84) &  38.70 (13.32) &  28.92 (08.34) \\
                       & GIN-MCD &  27.63 (01.13) &    25.00 &            30.83 (08.44) &  40.71 (00.17) &  46.04 (02.28) &  39.83 (12.66) &  35.64 (08.35) \\
\midrule
\textsc{Cora}          & GCN     &  86.00 (01.25) &    14.29 &             4.13 (00.87) &   6.29 (00.85) &   5.87 (01.34) &   4.88 (00.91) &   5.31 (01.07) \\
                       & GCN-MCD &  86.00 (01.02) &    14.29 &             9.47 (00.96) &   4.91 (01.24) &   4.74 (00.77) &   4.14 (00.76) &   4.37 (00.91) \\
                       & GAT     &  86.31 (01.07) &    14.29 &            40.62 (01.03) &   5.36 (00.95) &   3.19 (00.60) &   3.83 (00.78) &   4.38 (00.61) \\
                       & GAT-MCD &  85.30 (01.07) &    14.29 &            37.64 (01.87) &   3.44 (00.76) &   3.13 (00.62) &   4.17 (00.68) &   4.34 (00.98) \\
                       & GIN     &  84.61 (00.80) &    14.29 &             8.56 (00.76) &   5.11 (01.03) &   5.42 (00.65) &   4.97 (01.38) &   6.11 (00.83) \\
                       & GIN-MCD &  83.44 (01.49) &    14.29 &             4.01 (00.67) &   7.21 (01.17) &   4.87 (01.36) &   4.95 (01.20) &   5.22 (00.94) \\
\midrule
\textsc{PubMed}        & GCN     &  87.63 (00.42) &    33.33 &             1.65 (00.30) &   1.69 (00.24) &   1.43 (00.33) &   1.17 (00.26) &   1.19 (00.22) \\
                       & GCN-MCD &  87.67 (00.50) &    33.33 &             4.05 (00.42) &   2.08 (00.43) &   1.39 (00.37) &   1.29 (00.21) &   1.04 (00.23) \\
                       & GAT     &  86.92 (00.50) &    33.33 &             3.41 (00.50) &   2.00 (00.33) &   1.87 (00.28) &   1.49 (00.22) &   1.52 (00.43) \\
                       & GAT-MCD &  87.03 (00.45) &    33.33 &             5.14 (00.37) &   1.76 (00.26) &   1.99 (00.26) &   1.57 (00.32) &   1.70 (00.39) \\
                       & GIN     &  86.66 (00.42) &    33.33 &             4.62 (00.32) &   1.80 (00.50) &   1.69 (00.26) &   1.56 (00.34) &   1.78 (00.37) \\
                       & GIN-MCD &  86.30 (00.59) &    33.33 &             3.78 (00.39) &   1.82 (00.37) &   1.79 (00.49) &   2.00 (00.31) &   2.07 (00.36) \\
\midrule
\textsc{CiteSeer}      & GCN     &  74.39 (01.76) &    16.67 &            18.56 (02.01) &   4.22 (00.79) &   5.39 (01.18) &   4.40 (01.40) &   5.13 (00.96) \\
                       & GCN-MCD &  74.90 (01.32) &    16.67 &            24.23 (02.12) &   5.46 (01.62) &   5.59 (01.64) &   5.48 (00.85) &   5.27 (01.18) \\
                       & GAT     &  73.57 (01.62) &    16.67 &            28.55 (01.80) &   5.91 (01.43) &   7.16 (01.62) &   5.59 (01.85) &   5.53 (01.33) \\
                       & GAT-MCD &  72.05 (01.82) &    16.67 &            30.81 (01.39) &   4.87 (01.33) &   5.19 (01.56) &   5.31 (01.47) &   6.77 (01.17) \\
                       & GIN     &  62.85 (01.70) &    16.67 &            12.46 (01.67) &   4.45 (01.09) &   6.47 (01.29) &   7.22 (01.46) &   6.88 (02.11) \\
                       & GIN-MCD &  62.88 (01.24) &    16.67 &             5.43 (01.62) &   5.52 (01.70) &   6.06 (01.99) &   5.64 (00.61) &   6.55 (01.93) \\
\midrule
\textsc{Cora-Full}     & GCN     &  69.62 (00.87) &     1.43 &             7.34 (01.22) &   7.51 (01.06) &   7.13 (00.71) &   4.38 (00.67) &   4.96 (00.91) \\
                       & GCN-MCD &  69.49 (00.91) &     1.43 &             5.08 (00.78) &   6.98 (00.73) &   4.90 (00.83) &   3.73 (00.61) &   3.20 (00.73) \\
                       & GAT     &  68.81 (01.17) &     1.43 &             4.30 (00.68) &   6.41 (00.87) &   5.99 (00.95) &   3.22 (00.82) &   3.99 (00.96) \\
                       & GAT-MCD &  69.39 (00.91) &     1.43 &             3.16 (00.56) &   6.40 (00.90) &   5.39 (00.63) &   3.63 (00.69) &   3.37 (00.67) \\
                       & GIN     &  66.71 (00.63) &     1.43 &             5.64 (01.25) &   5.77 (00.97) &   6.60 (00.53) &   6.57 (01.14) &   8.69 (00.63) \\
                       & GIN-MCD &  65.84 (01.14) &     1.43 &            25.38 (01.53) &   6.56 (00.62) &   2.58 (00.73) &   4.70 (00.72) &   4.91 (00.70) \\
\midrule
\textsc{Facebook}      & GCN     &  70.15 (01.44) &    50.00 &             6.16 (01.55) &  12.59 (01.03) &   7.55 (00.86) &   5.49 (00.87) &   5.64 (01.13) \\
                       & GCN-MCD &  69.49 (01.08) &    50.00 &             5.31 (00.80) &  12.95 (01.18) &   5.73 (01.42) &   4.55 (01.08) &   5.50 (01.01) \\
                       & GAT     &  67.19 (02.08) &    50.00 &             5.92 (01.57) &  15.99 (00.91) &   6.78 (01.09) &   4.02 (00.76) &   4.82 (01.07) \\
                       & GAT-MCD &  68.62 (01.23) &    50.00 &             6.28 (00.93) &   8.92 (01.13) &   6.39 (01.04) &   5.75 (01.26) &   5.35 (01.15) \\
                       & GIN     &  65.11 (01.24) &    50.00 &             5.61 (00.79) &  11.85 (02.08) &   6.14 (01.26) &   5.65 (01.48) &   5.64 (01.83) \\
                       & GIN-MCD &  65.64 (01.68) &    50.00 &             4.78 (00.98) &  11.88 (01.57) &   6.37 (00.81) &   5.48 (01.58) &   6.21 (01.70) \\
\midrule
\textsc{Amazon-Cmp.}   & GCN     &  91.51 (00.54) &    10.00 &             3.40 (00.47) &   2.79 (00.28) &   2.49 (00.44) &   1.74 (00.43) &   1.95 (00.43) \\
                       & GCN-MCD &  91.51 (00.60) &    10.00 &             3.67 (00.64) &   1.87 (00.48) &   2.14 (00.62) &   1.58 (00.39) &   1.96 (00.33) \\
                       & GAT     &  93.15 (00.46) &    10.00 &             1.68 (00.34) &   2.37 (00.59) &   2.35 (00.26) &   1.75 (00.40) &   2.61 (00.48) \\
                       & GAT-MCD &  93.20 (00.26) &    10.00 &             1.95 (00.32) &   1.93 (00.34) &   2.12 (00.30) &   1.42 (00.41) &   2.62 (00.47) \\
                       & GIN     &  92.11 (00.53) &    10.00 &             2.20 (00.46) &   4.02 (00.52) &   2.14 (00.34) &   1.87 (00.34) &   2.36 (00.46) \\
                       & GIN-MCD &  92.08 (00.53) &    10.00 &             6.67 (00.51) &   3.18 (00.54) &   2.03 (00.30) &   1.89 (00.51) &   2.05 (00.37) \\
\midrule
\textsc{Amazon-Ph.}    & GCN     &  93.61 (00.69) &    12.50 &             1.49 (00.43) &   2.29 (00.29) &   1.99 (00.36) &   1.82 (00.35) &   1.36 (00.47) \\
                       & GCN-MCD &  93.68 (00.86) &    12.50 &             1.31 (00.37) &   1.91 (00.56) &   1.67 (00.58) &   1.83 (00.22) &   1.40 (00.29) \\
                       & GAT     &  93.38 (00.77) &    12.50 &             2.79 (00.56) &   1.57 (00.36) &   1.74 (00.34) &   1.75 (00.49) &   1.44 (00.51) \\
                       & GAT-MCD &  93.04 (00.47) &    12.50 &             2.34 (00.43) &   2.09 (00.41) &   1.97 (00.43) &   1.61 (00.34) &   1.67 (00.35) \\
                       & GIN     &  93.96 (00.59) &    12.50 &             2.07 (00.31) &   2.02 (00.36) &   1.99 (00.37) &   1.47 (00.49) &   2.31 (00.67) \\
                       & GIN-MCD &  93.87 (00.61) &    12.50 &             1.65 (00.48) &   2.53 (00.25) &   1.80 (00.55) &   1.61 (00.43) &   2.51 (00.55) \\
\bottomrule
\end{tabular}
}
\caption{
    \textbf{(\(\text{ECE}_{\geq 50}\), Accuracy, Balanced)}
    Accuracy and \(\text{ECE}_{\geq 50}\) results for all models in all dataset, as well as \(\text{ECE}_{\geq 50}\) results for calibration methods.
    For reference, accuracy of random model (1 / number of classes) is also provided.
    For each row, the best model (over all tested hyperparameter configurations), 
    was chosen using validation accuracy. Metrics shown in the table are balanced.
}
\label{tab:results-ece-50-accuracy-balanced}
\end{table*}

\begin{table*}
\centering
\resizebox{\textwidth}{!}{
\begin{tabular}{llrrrrrrr}

\toprule
\multicolumn{9}{c}{Early Stopping using: Validation Loss, Test Metrics: Not Balanced} \\
\midrule
        &       & \multicolumn{2}{c}{ Accuracy (\%) }     &                                 \multicolumn{5}{c}{\(\text{ECE}_{\geq 50}\) (\%)} \\
\cmidrule(lr){3-4} \cmidrule(lr){5-9}
Dataset & Model & \multicolumn{1}{c}{Original} & \multicolumn{1}{c}{Random} & \multicolumn{1}{c}{Original}  & \multicolumn{1}{c}{Hist. Bin.}  & \multicolumn{1}{c}{Isotonic} & \multicolumn{1}{c}{TS w/ NLL} & \multicolumn{1}{c}{TS w/ Brier}  \\
\midrule
\textsc{Friendster}    & GCN     &  31.97 (00.60) &    25.00 &            39.44 (03.39) &   3.21 (00.51) &  25.69 (01.36) &  41.17 (02.86) &  39.72 (04.25) \\
                       & GCN-MCD &  31.17 (00.34) &    25.00 &            39.44 (08.06) &   3.07 (00.64) &  23.09 (01.27) &  37.63 (03.22) &  37.46 (01.84) \\
                       & GAT     &  24.67 (00.61) &    25.00 &            51.68 (02.81) &   2.05 (00.34) &  48.50 (01.69) &  53.40 (05.07) &  52.41 (04.87) \\
                       & GAT-MCD &  27.83 (00.51) &    25.00 &            53.27 (02.86) &   1.66 (00.40) &  43.17 (02.32) &  51.06 (04.46) &  53.79 (03.88) \\
                       & GIN     &  23.75 (00.39) &    25.00 &            36.50 (09.63) &   1.35 (00.47) &  45.21 (01.58) &  46.74 (07.32) &  45.49 (06.07) \\
                       & GIN-MCD &  24.23 (00.58) &    25.00 &            50.40 (06.66) &   1.32 (00.71) &  49.40 (01.63) &  45.15 (08.83) &  47.79 (04.04) \\
\midrule
\textsc{Cora}          & GCN     &  86.19 (01.12) &    14.29 &             4.00 (00.85) &   5.69 (01.03) &   6.12 (00.94) &   4.54 (00.92) &   5.19 (00.88) \\
                       & GCN-MCD &  85.30 (00.90) &    14.29 &             8.96 (00.79) &   4.80 (00.71) &   5.18 (00.52) &   3.72 (00.72) &   4.17 (00.94) \\
                       & GAT     &  85.46 (01.14) &    14.29 &             3.68 (00.69) &   4.53 (01.04) &   5.29 (00.78) &   4.56 (00.52) &   3.35 (00.95) \\
                       & GAT-MCD &  84.81 (01.01) &    14.29 &             3.36 (00.63) &   5.87 (01.30) &   5.97 (01.15) &   4.50 (00.93) &   3.21 (00.75) \\
                       & GIN     &  83.64 (00.72) &    14.29 &             5.53 (01.10) &   5.64 (00.91) &   5.23 (01.06) &   4.89 (01.09) &   4.37 (00.79) \\
                       & GIN-MCD &  84.14 (01.63) &    14.29 &             4.43 (00.68) &   5.11 (01.14) &   5.07 (00.78) &   5.08 (00.77) &   4.95 (00.86) \\
\midrule
\textsc{PubMed}        & GCN     &  87.96 (00.45) &    33.33 &             1.67 (00.38) &   1.47 (00.44) &   1.30 (00.26) &   1.58 (00.28) &   1.55 (00.33) \\
                       & GCN-MCD &  87.98 (00.35) &    33.33 &             3.26 (00.46) &   1.37 (00.17) &   1.52 (00.44) &   1.22 (00.24) &   1.39 (00.32) \\
                       & GAT     &  87.71 (00.49) &    33.33 &             1.96 (00.37) &   2.11 (00.40) &   1.66 (00.37) &   1.58 (00.58) &   1.29 (00.45) \\
                       & GAT-MCD &  87.93 (00.57) &    33.33 &             1.49 (00.31) &   1.85 (00.33) &   1.83 (00.25) &   1.61 (00.35) &   1.39 (00.30) \\
                       & GIN     &  87.31 (00.69) &    33.33 &             1.23 (00.34) &   1.40 (00.33) &   1.64 (00.39) &   1.16 (00.42) &   1.12 (00.31) \\
                       & GIN-MCD &  87.23 (00.43) &    33.33 &             1.27 (00.18) &   1.58 (00.27) &   1.63 (00.34) &   1.11 (00.21) &   1.48 (00.24) \\
\midrule
\textsc{CiteSeer}      & GCN     &  76.92 (01.15) &    16.67 &             4.64 (01.23) &   4.65 (00.84) &   4.85 (01.11) &   3.85 (01.03) &   4.48 (01.08) \\
                       & GCN-MCD &  77.57 (01.06) &    16.67 &            11.68 (00.83) &   4.02 (01.10) &   5.23 (01.05) &   3.73 (00.84) &   4.47 (00.99) \\
                       & GAT     &  75.50 (01.46) &    16.67 &             9.71 (01.26) &   4.14 (00.91) &   5.01 (01.46) &   4.14 (00.61) &   3.87 (00.99) \\
                       & GAT-MCD &  74.66 (01.67) &    16.67 &             9.96 (01.48) &   4.09 (01.23) &   5.08 (00.65) &   3.36 (00.88) &   3.64 (00.56) \\
                       & GIN     &  69.24 (01.42) &    16.67 &             6.70 (01.53) &   6.07 (00.80) &   5.86 (01.27) &   6.23 (01.43) &   5.89 (01.00) \\
                       & GIN-MCD &  68.86 (01.47) &    16.67 &             6.66 (01.62) &   5.63 (01.36) &   5.39 (01.32) &   5.96 (01.18) &   6.26 (01.39) \\
\midrule
\textsc{Cora-Full}     & GCN     &  69.87 (00.61) &     1.43 &             3.65 (00.42) &   4.93 (00.62) &   5.05 (00.49) &   3.60 (00.37) &   3.68 (00.39) \\
                       & GCN-MCD &  69.93 (00.74) &     1.43 &            11.24 (00.88) &   2.40 (00.29) &   2.87 (00.53) &   2.96 (00.53) &   2.95 (00.57) \\
                       & GAT     &  69.67 (00.46) &     1.43 &             3.08 (00.29) &   4.70 (00.61) &   4.54 (00.45) &   3.22 (00.62) &   3.14 (00.40) \\
                       & GAT-MCD &  69.14 (00.72) &     1.43 &             2.15 (00.50) &   4.55 (00.71) &   4.01 (00.40) &   2.13 (00.49) &   2.53 (00.48) \\
                       & GIN     &  63.50 (00.85) &     1.43 &             4.46 (00.82) &   4.81 (00.64) &   3.23 (00.58) &   6.60 (00.60) &   4.24 (00.85) \\
                       & GIN-MCD &  63.30 (00.60) &     1.43 &            20.16 (00.64) &   4.30 (00.79) &   2.70 (00.38) &   4.13 (00.63) &   2.37 (00.52) \\
\midrule
\textsc{Facebook}      & GCN     &  68.87 (01.38) &    50.00 &             5.88 (01.18) &   6.31 (01.47) &   9.20 (01.26) &   6.01 (00.95) &   4.94 (01.03) \\
                       & GCN-MCD &  69.30 (01.27) &    50.00 &             6.37 (01.15) &   5.51 (01.04) &   8.16 (00.75) &   5.99 (00.83) &   5.86 (00.80) \\
                       & GAT     &  65.02 (01.83) &    50.00 &             4.43 (01.03) &   4.55 (01.04) &   7.68 (01.38) &   4.29 (01.17) &   4.19 (01.04) \\
                       & GAT-MCD &  67.07 (01.81) &    50.00 &             5.39 (01.15) &   2.98 (01.02) &   7.56 (01.36) &   4.57 (01.02) &   3.79 (00.83) \\
                       & GIN     &  66.33 (01.93) &    50.00 &             4.83 (01.87) &   3.98 (00.97) &   8.85 (00.90) &   4.40 (00.91) &   4.51 (01.11) \\
                       & GIN-MCD &  66.06 (01.91) &    50.00 &             3.77 (00.95) &   3.40 (00.96) &   9.67 (01.27) &   4.60 (00.80) &   5.01 (00.97) \\
\midrule
\textsc{Amazon-Cmp.}   & GCN     &  87.59 (00.65) &    10.00 &             3.41 (00.52) &   1.56 (00.36) &   3.71 (00.44) &   2.44 (00.42) &   3.26 (00.38) \\
                       & GCN-MCD &  87.80 (00.30) &    10.00 &             3.73 (00.50) &   1.98 (00.32) &   3.63 (00.64) &   2.18 (00.45) &   3.13 (00.50) \\
                       & GAT     &  89.42 (00.52) &    10.00 &             1.65 (00.49) &   2.00 (00.35) &   3.75 (00.38) &   2.07 (00.52) &   1.92 (00.35) \\
                       & GAT-MCD &  89.63 (00.64) &    10.00 &             1.38 (00.51) &   2.19 (00.39) &   3.35 (00.65) &   1.80 (00.53) &   1.69 (00.34) \\
                       & GIN     &  89.28 (00.68) &    10.00 &             2.78 (00.48) &   1.54 (00.44) &   4.18 (00.40) &   2.74 (00.19) &   2.78 (00.45) \\
                       & GIN-MCD &  89.17 (00.65) &    10.00 &             2.35 (00.31) &   1.80 (00.22) &   4.24 (00.37) &   2.55 (00.39) &   2.54 (00.36) \\
\midrule
\textsc{Amazon-Ph.}    & GCN     &  93.36 (00.61) &    12.50 &             1.59 (00.42) &   1.93 (00.32) &   1.80 (00.40) &   1.67 (00.40) &   1.76 (00.41) \\
                       & GCN-MCD &  93.01 (00.69) &    12.50 &             1.52 (00.25) &   1.52 (00.32) &   1.49 (00.31) &   1.61 (00.46) &   1.74 (00.52) \\
                       & GAT     &  93.27 (00.57) &    12.50 &             1.37 (00.21) &   1.88 (00.58) &   1.51 (00.29) &   1.37 (00.38) &   1.57 (00.47) \\
                       & GAT-MCD &  93.38 (00.46) &    12.50 &             1.62 (00.32) &   1.84 (00.43) &   1.44 (00.22) &   1.40 (00.31) &   1.42 (00.27) \\
                       & GIN     &  92.53 (00.75) &    12.50 &             2.83 (00.68) &   2.62 (00.65) &   2.71 (00.50) &   1.46 (00.32) &   3.34 (00.43) \\
                       & GIN-MCD &  92.20 (00.52) &    12.50 &             1.67 (00.28) &   2.02 (00.59) &   2.48 (00.53) &   1.83 (00.39) &   2.50 (00.42) \\
\bottomrule
\end{tabular}
}
\caption{
    \textbf{(\(\text{ECE}_{\geq 50}\), Loss, Not Balanced)}
    Accuracy and \(\text{ECE}_{\geq 50}\) results for all models in all dataset, as well as \(\text{ECE}_{\geq 50}\) results for calibration methods.
    For reference, accuracy of random model (1 / number of classes) is also provided.
    For each row, the best model (over all tested hyperparameter configurations), 
    was chosen using validation loss. Metrics shown in the table are not balanced.
}
\label{tab:results-ece-50-loss-not-balanced}
\end{table*}

\begin{table*}
\centering
\resizebox{\textwidth}{!}{
\begin{tabular}{llrrrrrrr}

\toprule
\multicolumn{9}{c}{Early Stopping using: Validation Los, Test Metrics: Balanced} \\
\midrule
        &       & \multicolumn{2}{c}{ Accuracy (\%) }     &                                 \multicolumn{5}{c}{\(\text{ECE}_{\geq 50}\) (\%)} \\
\cmidrule(lr){3-4} \cmidrule(lr){5-9}
Dataset & Model & \multicolumn{1}{c}{Original} & \multicolumn{1}{c}{Random} & \multicolumn{1}{c}{Original}  & \multicolumn{1}{c}{Hist. Bin.}  & \multicolumn{1}{c}{Isotonic} & \multicolumn{1}{c}{TS w/ NLL} & \multicolumn{1}{c}{TS w/ Brier} \\
\midrule
\textsc{Friendster}    & GCN     &  35.11 (01.35) &    25.00 &            36.86 (03.79) &  39.93 (00.26) &  14.34 (02.36) &  30.40 (05.62) &  32.71 (02.94) \\
                       & GCN-MCD &  33.31 (01.21) &    25.00 &            32.19 (08.26) &  39.99 (00.12) &  14.98 (04.16) &  17.95 (07.01) &  20.07 (07.31) \\
                       & GAT     &  30.71 (01.87) &    25.00 &            43.96 (02.14) &  40.43 (00.19) &  40.24 (04.01) &  47.90 (07.28) &  45.95 (06.57) \\
                       & GAT-MCD &  28.81 (01.44) &    25.00 &            48.14 (07.09) &  40.55 (00.19) &  35.68 (06.44) &  45.77 (05.97) &  48.89 (06.45) \\
                       & GIN     &  26.82 (00.91) &    25.00 &            36.37 (08.71) &  40.77 (00.14) &  43.85 (02.13) &  42.14 (13.11) &  32.86 (13.42) \\
                       & GIN-MCD &  26.90 (00.82) &    25.00 &            47.85 (08.33) &  40.61 (00.13) &  47.57 (01.82) &  37.64 (12.65) &  33.59 (10.89) \\
\midrule
\textsc{Cora}          & GCN     &  86.73 (01.81) &    14.29 &             3.40 (00.76) &   6.77 (00.95) &   6.30 (00.89) &   4.91 (01.33) &   5.53 (01.18) \\
                       & GCN-MCD &  85.50 (01.10) &    14.29 &             8.86 (01.52) &   6.60 (01.14) &   5.75 (00.76) &   3.62 (00.85) &   4.43 (00.98) \\
                       & GAT     &  84.24 (00.99) &    14.29 &             4.21 (00.77) &   5.35 (01.05) &   5.73 (00.67) &   4.76 (00.86) &   4.79 (00.98) \\
                       & GAT-MCD &  85.10 (01.67) &    14.29 &             3.71 (00.83) &   6.47 (00.81) &   5.32 (00.87) &   4.44 (01.00) &   3.94 (01.22) \\
                       & GIN     &  83.76 (01.19) &    14.29 &             6.45 (01.47) &   6.07 (00.53) &   4.93 (00.98) &   5.34 (00.80) &   5.12 (00.90) \\
                       & GIN-MCD &  83.00 (01.00) &    14.29 &             4.43 (00.80) &   4.31 (01.08) &   4.85 (00.68) &   5.07 (00.82) &   5.47 (01.17) \\
\midrule
\textsc{PubMed}        & GCN     &  88.73 (00.57) &    33.33 &             1.62 (00.28) &   1.42 (00.35) &   1.35 (00.33) &   1.55 (00.34) &   1.45 (00.42) \\
                       & GCN-MCD &  88.47 (00.44) &    33.33 &             3.62 (00.21) &   1.60 (00.30) &   1.32 (00.33) &   1.33 (00.26) &   1.37 (00.26) \\
                       & GAT     &  88.19 (00.36) &    33.33 &             1.66 (00.28) &   2.01 (00.29) &   1.27 (00.30) &   1.46 (00.23) &   1.33 (00.35) \\
                       & GAT-MCD &  87.72 (00.38) &    33.33 &             1.43 (00.19) &   1.99 (00.40) &   1.24 (00.27) &   1.46 (00.31) &   1.25 (00.20) \\
                       & GIN     &  87.38 (00.54) &    33.33 &             1.22 (00.32) &   1.77 (00.33) &   1.43 (00.39) &   1.13 (00.32) &   1.26 (00.20) \\
                       & GIN-MCD &  87.31 (00.60) &    33.33 &             1.20 (00.39) &   2.26 (00.20) &   1.45 (00.25) &   1.21 (00.33) &   1.38 (00.32) \\
\midrule
\textsc{CiteSeer}      & GCN     &  74.53 (02.07) &    16.67 &             5.68 (01.49) &   6.03 (01.38) &   5.33 (01.02) &   4.80 (01.02) &   5.85 (01.09) \\
                       & GCN-MCD &  75.03 (02.35) &    16.67 &            11.13 (01.76) &   6.09 (01.60) &   5.82 (01.90) &   5.15 (01.20) &   4.80 (01.47) \\
                       & GAT     &  74.72 (01.72) &    16.67 &             8.72 (01.92) &   5.90 (01.24) &   6.03 (01.38) &   4.71 (00.92) &   4.34 (01.00) \\
                       & GAT-MCD &  74.16 (01.64) &    16.67 &             9.12 (01.52) &   5.92 (00.99) &   5.95 (01.26) &   4.35 (00.94) &   4.55 (00.95) \\
                       & GIN     &  69.49 (01.33) &    16.67 &             7.33 (01.62) &   5.02 (01.38) &   6.60 (01.74) &   6.44 (01.42) &   7.36 (01.09) \\
                       & GIN-MCD &  68.66 (01.33) &    16.67 &             7.44 (01.27) &   4.98 (01.42) &   6.86 (01.50) &   6.88 (01.04) &   6.66 (01.52) \\
\midrule
\textsc{Cora-Full}     & GCN     &  68.98 (01.05) &     1.43 &             4.02 (00.72) &   6.46 (00.81) &   6.80 (01.05) &   3.74 (00.64) &   4.03 (00.66) \\
                       & GCN-MCD &  68.98 (00.90) &     1.43 &            11.26 (01.24) &   5.84 (00.79) &   3.25 (00.67) &   4.19 (00.53) &   3.88 (00.86) \\
                       & GAT     &  68.85 (00.93) &     1.43 &             4.44 (00.69) &   6.86 (01.34) &   5.23 (00.98) &   3.07 (00.75) &   4.06 (00.63) \\
                       & GAT-MCD &  68.89 (01.24) &     1.43 &             2.98 (00.75) &   6.86 (00.60) &   5.16 (01.03) &   2.95 (00.95) &   3.32 (00.77) \\
                       & GIN     &  62.45 (01.13) &     1.43 &             3.73 (00.83) &   4.99 (01.14) &   4.18 (00.73) &   5.32 (00.84) &   5.00 (00.99) \\
                       & GIN-MCD &  62.76 (00.84) &     1.43 &            18.83 (01.33) &   3.55 (00.98) &   3.45 (00.63) &   3.92 (00.72) &   3.59 (00.62) \\
\midrule
\textsc{Facebook}      & GCN     &  70.08 (01.37) &    50.00 &             5.45 (01.11) &   9.79 (00.56) &   7.30 (00.94) &   5.06 (00.61) &   5.84 (01.02) \\
                       & GCN-MCD &  69.62 (01.80) &    50.00 &             5.62 (01.17) &   8.77 (01.14) &   6.99 (01.20) &   6.00 (01.33) &   5.87 (01.07) \\
                       & GAT     &  66.15 (01.42) &    50.00 &             3.38 (00.85) &   8.45 (00.97) &   4.39 (01.18) &   3.68 (00.94) &   3.87 (01.43) \\
                       & GAT-MCD &  66.57 (01.62) &    50.00 &             4.37 (01.64) &   9.88 (01.11) &   4.12 (01.22) &   4.35 (01.13) &   4.48 (00.93) \\
                       & GIN     &  64.93 (01.45) &    50.00 &             5.52 (01.17) &  11.09 (01.10) &   6.31 (00.84) &   4.92 (00.62) &   5.29 (01.04) \\
                       & GIN-MCD &  64.78 (01.34) &    50.00 &             4.46 (00.84) &  12.23 (00.73) &   6.01 (01.14) &   5.19 (01.28) &   5.03 (01.21) \\
\midrule
\textsc{Amazon-Cmp.}   & GCN     &  91.31 (00.64) &    10.00 &             3.06 (00.31) &   1.80 (00.57) &   2.39 (00.42) &   1.77 (00.33) &   2.46 (00.39) \\
                       & GCN-MCD &  91.60 (00.49) &    10.00 &             3.54 (00.26) &   2.00 (00.56) &   2.21 (00.30) &   1.56 (00.32) &   2.16 (00.53) \\
                       & GAT     &  92.87 (00.60) &    10.00 &             1.49 (00.41) &   2.15 (00.54) &   2.51 (00.38) &   1.94 (00.53) &   1.77 (00.50) \\
                       & GAT-MCD &  92.54 (00.65) &    10.00 &             1.18 (00.37) &   2.43 (00.43) &   2.41 (00.46) &   1.35 (00.22) &   1.51 (00.28) \\
                       & GIN     &  91.10 (00.48) &    10.00 &             2.35 (00.55) &   2.95 (00.59) &   2.58 (00.57) &   2.22 (00.25) &   2.11 (00.46) \\
                       & GIN-MCD &  91.33 (00.71) &    10.00 &             1.99 (00.48) &   2.23 (00.41) &   2.48 (00.43) &   2.17 (00.52) &   2.33 (00.47) \\
\midrule
\textsc{Amazon-Ph.}    & GCN     &  94.24 (00.58) &    12.50 &             1.66 (00.37) &   2.33 (00.57) &   1.57 (00.27) &   1.77 (00.41) &   1.60 (00.36) \\
                       & GCN-MCD &  94.12 (00.58) &    12.50 &             1.86 (00.27) &   2.31 (00.49) &   1.70 (00.27) &   1.85 (00.34) &   1.46 (00.37) \\
                       & GAT     &  93.91 (00.47) &    12.50 &             1.40 (00.39) &   2.59 (00.56) &   1.59 (00.40) &   1.71 (00.28) &   1.79 (00.57) \\
                       & GAT-MCD &  93.33 (00.71) &    12.50 &             1.91 (00.62) &   2.66 (00.69) &   1.57 (00.48) &   1.70 (00.42) &   1.77 (00.38) \\
                       & GIN     &  93.31 (00.79) &    12.50 &             2.70 (00.52) &   3.21 (00.67) &   2.95 (00.54) &   1.59 (00.45) &   2.93 (00.28) \\
                       & GIN-MCD &  93.04 (00.40) &    12.50 &             1.97 (00.51) &   2.69 (00.76) &   2.38 (00.77) &   2.25 (00.38) &   2.65 (00.50) \\
\bottomrule
\end{tabular}
}
\caption{
    \textbf{(\(\text{ECE}_{\geq 50}\), Loss, Balanced)}
    Accuracy and \(\text{ECE}_{\geq 50}\) results for all models in all dataset, as well as \(\text{ECE}_{\geq 50}\) results for calibration methods.
    For reference, accuracy of random model (1 / number of classes) is also provided.
    For each row, the best model (over all tested hyperparameter configurations), 
    was chosen using validation loss. Metrics shown in the table are balanced.
}
\label{tab:results-ece-50-loss-balanced}
\end{table*}


%% file: ms.bbl
\begin{thebibliography}{44}
\providecommand{\natexlab}[1]{#1}
\providecommand{\url}[1]{\texttt{#1}}
\expandafter\ifx\csname urlstyle\endcsname\relax
  \providecommand{\doi}[1]{doi: #1}\else
  \providecommand{\doi}{doi: \begingroup \urlstyle{rm}\Url}\fi

\bibitem[Battaglia et~al.(2018)Battaglia, Hamrick, Bapst, {Sanchez-Gonzalez},
  Zambaldi, Malinowski, Tacchetti, Raposo, Santoro, Faulkner, Gulcehre, Song,
  Ballard, Gilmer, Dahl, Vaswani, Allen, Nash, Langston, Dyer, Heess, Wierstra,
  Kohli, Botvinick, Vinyals, Li, and Pascanu]{Battaglia2018}
Battaglia, P.~W., Hamrick, J.~B., Bapst, V., {Sanchez-Gonzalez}, A., Zambaldi,
  V., Malinowski, M., Tacchetti, A., Raposo, D., Santoro, A., Faulkner, R.,
  Gulcehre, C., Song, F., Ballard, A., Gilmer, J., Dahl, G., Vaswani, A.,
  Allen, K., Nash, C., Langston, V., Dyer, C., Heess, N., Wierstra, D., Kohli,
  P., Botvinick, M., Vinyals, O., Li, Y., and Pascanu, R.
\newblock Relational inductive biases, deep learning, and graph networks.
\newblock \emph{arXiv:1806.01261 [cs, stat]}, 2018.
\newblock URL \url{http://arxiv.org/abs/1806.01261}.

\bibitem[Blattenberger \& Lad(1985)Blattenberger and Lad]{Blattenberger1985}
Blattenberger, G. and Lad, F.
\newblock Separating the {{Brier Score}} into {{Calibration}} and {{Refinement
  Components}}: {{A Graphical Exposition}}.
\newblock \emph{The American Statistician}, 39\penalty0 (1):\penalty0 26--32,
  February 1985.
\newblock ISSN 0003-1305.
\newblock \doi{10.1080/00031305.1985.10479382}.

\bibitem[Box(1980)]{box1980sampling}
Box, G.~E.
\newblock Sampling and bayes' inference in scientific modelling and robustness.
\newblock \emph{Journal of the Royal Statistical Society: Series A (General)},
  143\penalty0 (4):\penalty0 383--404, 1980.

\bibitem[Brier(1950)]{Brier1950}
Brier, G.~W.
\newblock Verification of forecasts expressed in terms of probability.
\newblock \emph{Monthly Weather Review}, 78\penalty0 (1):\penalty0 1--3,
  January 1950.
\newblock ISSN 0027-0644.
\newblock \doi{10.1175/1520-0493(1950)078<0001:VOFEIT>2.0.CO;2}.

\bibitem[Card et~al.(2019)Card, Zhang, and Smith]{Card2019}
Card, D., Zhang, M., and Smith, N.~A.
\newblock Deep {{Weighted Averaging Classifiers}}.
\newblock In \emph{Proceedings of the {{Conference}} on {{Fairness}},
  {{Accountability}}, and {{Transparency}}}, pp.\  369--378, New York, NY, USA,
  2019. {ACM}.
\newblock ISBN 978-1-4503-6125-5.
\newblock \doi{10.1145/3287560.3287595}.

\bibitem[DeGroot \& Fienberg(1983)DeGroot and Fienberg]{DeGroot1983}
DeGroot, M.~H. and Fienberg, S.~E.
\newblock The {{Comparison}} and {{Evaluation}} of {{Forecasters}}.
\newblock \emph{Journal of the Royal Statistical Society. Series D (The
  Statistician)}, 32\penalty0 (1/2):\penalty0 12--22, 1983.
\newblock ISSN 0039-0526.
\newblock \doi{10.2307/2987588}.

\bibitem[Duvenaud et~al.(2015)Duvenaud, Maclaurin, Iparraguirre, Bombarell,
  Hirzel, {Aspuru-Guzik}, and Adams]{Duvenaud2015}
Duvenaud, D.~K., Maclaurin, D., Iparraguirre, J., Bombarell, R., Hirzel, T.,
  {Aspuru-Guzik}, A., and Adams, R.~P.
\newblock Convolutional {{Networks}} on {{Graphs}} for {{Learning Molecular
  Fingerprints}}.
\newblock In Cortes, C., Lawrence, N.~D., Lee, D.~D., Sugiyama, M., and
  Garnett, R. (eds.), \emph{Advances in {{Neural Information Processing
  Systems}} 28}, pp.\  2224--2232. {Curran Associates, Inc.}, 2015.

\bibitem[Fey \& Lenssen(2019)Fey and Lenssen]{Fey-Lenssen-2019}
Fey, M. and Lenssen, J.~E.
\newblock Fast graph representation learning with {PyTorch Geometric}.
\newblock In \emph{ICLR Workshop on Representation Learning on Graphs and
  Manifolds}, 2019.

\bibitem[Gal \& Ghahramani(2016)Gal and Ghahramani]{Gal2016}
Gal, Y. and Ghahramani, Z.
\newblock Dropout as a {{Bayesian Approximation}}: {{Representing Model
  Uncertainty}} in {{Deep Learning}}.
\newblock In \emph{Proceedings of {{The}} 33rd {{International Conference}} on
  {{Machine Learning}}, {{PMLR}}}, volume~48, pp.\  1050--1059, New York, NY,
  USA, June 2016. {JMLR.org}.

\bibitem[Gao et~al.(2017)Gao, Parameswaran, and Peng]{Gao2017}
Gao, Y., Parameswaran, A., and Peng, J.
\newblock On the {{Interpretability}} of {{Conditional Probability Estimates}}
  in the {{Agnostic Setting}}.
\newblock In \emph{Artificial {{Intelligence}} and {{Statistics}}}, pp.\
  1367--1374, April 2017.

\bibitem[Gilmer et~al.(2017)Gilmer, Schoenholz, Riley, Vinyals, and
  Dahl]{Gilmer2017}
Gilmer, J., Schoenholz, S.~S., Riley, P.~F., Vinyals, O., and Dahl, G.~E.
\newblock Neural {{Message Passing}} for {{Quantum Chemistry}}.
\newblock In \emph{International {{Conference}} on {{Machine Learning}}}, pp.\
  1263--1272, July 2017.

\bibitem[Guo et~al.(2017)Guo, Pleiss, Sun, and Weinberger]{Guo2017}
Guo, C., Pleiss, G., Sun, Y., and Weinberger, K.~Q.
\newblock On {{Calibration}} of {{Modern Neural Networks}}.
\newblock In \emph{International {{Conference}} on {{Machine Learning}}}, pp.\
  1321--1330, July 2017.

\bibitem[Ioffe \& Szegedy(2015)Ioffe and Szegedy]{Ioffe2015}
Ioffe, S. and Szegedy, C.
\newblock Batch {{Normalization}}: {{Accelerating Deep Network Training}} by
  {{Reducing Internal Covariate Shift}}.
\newblock In \emph{International {{Conference}} on {{Machine Learning}}}, pp.\
  448--456, June 2015.

\bibitem[Jones et~al.(2001--)Jones, Oliphant, Peterson, et~al.]{SciPy}
Jones, E., Oliphant, T., Peterson, P., et~al.
\newblock {SciPy}: Open source scientific tools for {Python}, 2001--.
\newblock URL \url{http://www.scipy.org/}.
\newblock [Online; accessed <today>].

\bibitem[Kingma \& Ba(2014)Kingma and Ba]{Kingma2014}
Kingma, D.~P. and Ba, J.
\newblock Adam: {{A Method}} for {{Stochastic Optimization}}.
\newblock In \emph{International {{Conference}} on {{Learning
  Representations}}}, 2014.

\bibitem[Kipf \& Welling(2017)Kipf and Welling]{Kipf2017}
Kipf, T.~N. and Welling, M.
\newblock Semi-{{Supervised Classification}} with {{Graph Convolutional
  Networks}}.
\newblock In \emph{International {{Conference}} on {{Learning
  Representations}}}, 2017.

\bibitem[Kuleshov et~al.(2018)Kuleshov, Fenner, and Ermon]{Kuleshov2018}
Kuleshov, V., Fenner, N., and Ermon, S.
\newblock Accurate {{Uncertainties}} for {{Deep Learning Using Calibrated
  Regression}}.
\newblock In \emph{International {{Conference}} on {{Machine Learning}}}, pp.\
  2796--2804, July 2018.

\bibitem[Kumar et~al.(2018)Kumar, Sarawagi, and Jain]{Kumar2018}
Kumar, A., Sarawagi, S., and Jain, U.
\newblock Trainable {{Calibration Measures}} for {{Neural Networks}} from
  {{Kernel Mean Embeddings}}.
\newblock In \emph{International {{Conference}} on {{Machine Learning}}}, pp.\
  2805--2814, July 2018.

\bibitem[Laakso \& Taagepera(1979)Laakso and Taagepera]{Laakso1979}
Laakso, M. and Taagepera, R.
\newblock “effective” number of parties: a measure with application to west
  europe.
\newblock \emph{Comparative political studies}, 12\penalty0 (1):\penalty0
  3--27, 1979.

\bibitem[Lakshminarayanan et~al.(2017)Lakshminarayanan, Pritzel, and
  Blundell]{Lakshminarayanan2017}
Lakshminarayanan, B., Pritzel, A., and Blundell, C.
\newblock Simple and {{Scalable Predictive Uncertainty Estimation}} using
  {{Deep Ensembles}}.
\newblock In Guyon, I., Luxburg, U.~V., Bengio, S., Wallach, H., Fergus, R.,
  Vishwanathan, S., and Garnett, R. (eds.), \emph{Advances in {{Neural
  Information Processing Systems}} 30}, pp.\  6402--6413. {Curran Associates,
  Inc.}, 2017.

\bibitem[Leskovec \& Faloutsos(2006)Leskovec and Faloutsos]{Leskovec2006}
Leskovec, J. and Faloutsos, C.
\newblock Sampling from large graphs.
\newblock In \emph{Proceedings of the 12th {{ACM SIGKDD}} International
  Conference on {{Knowledge Discovery}} and {{Data Mining}}}, pp.\  631--636,
  2006.

\bibitem[Linderman et~al.(2019)Linderman, Rachh, Hoskins, Steinerberger, and
  Kluger]{Linderman2019}
Linderman, G.~C., Rachh, M., Hoskins, J.~G., Steinerberger, S., and Kluger, Y.
\newblock Fast interpolation-based t-{{SNE}} for improved visualization of
  single-cell {{RNA}}-seq data.
\newblock \emph{Nature Methods}, 16\penalty0 (3):\penalty0 243, March 2019.
\newblock ISSN 1548-7105.
\newblock \doi{10.1038/s41592-018-0308-4}.

\bibitem[Maaten \& Hinton(2008)Maaten and Hinton]{maaten2008}
Maaten, L. v.~d. and Hinton, G.
\newblock Visualizing data using t-sne.
\newblock \emph{Journal of machine learning research}, 9\penalty0
  (Nov):\penalty0 2579--2605, 2008.

\bibitem[Mozafari et~al.(2018)Mozafari, Gomes, Janny, and
  Gagn\'e]{Mozafari2018}
Mozafari, A.~S., Gomes, H.~S., Janny, S., and Gagn\'e, C.
\newblock A {{New Loss Function}} for {{Temperature Scaling}} to have {{Better
  Calibrated Deep Networks}}.
\newblock In \emph{{{NeurIPS}} - {{Workshop}} on {{Security}} in {{Machine
  Learning}}}, October 2018.

\bibitem[Murphy et~al.(2019)Murphy, Srinivasan, Rao, and Ribeiro]{Murphy2019a}
Murphy, R.~L., Srinivasan, B., Rao, V., and Ribeiro, B.
\newblock Janossy pooling: Learning deep permutation-invariant functions for
  variable-size inputs.
\newblock In \emph{ICLR}, 2019.

\bibitem[Naeini et~al.(2015)Naeini, Cooper, and Hauskrecht]{Naeini2015}
Naeini, M.~P., Cooper, G.~F., and Hauskrecht, M.
\newblock Obtaining {{Well Calibrated Probabilities Using Bayesian Binning}}.
\newblock In \emph{Proceedings of the 29th {{AAAI Conference}} on {{Artificial
  Intelligence}}}, volume 2015, pp.\  2901--2907, January 2015.

\bibitem[Nair \& Hinton(2010)Nair and Hinton]{Nair2010}
Nair, V. and Hinton, G.~E.
\newblock Rectified linear units improve restricted boltzmann machines.
\newblock In \emph{Proceedings of the 27th international conference on machine
  learning (ICML-10)}, pp.\  807--814, 2010.

\bibitem[{Niculescu-Mizil} \& Caruana(2005){Niculescu-Mizil} and
  Caruana]{Niculescu-Mizil2005}
{Niculescu-Mizil}, A. and Caruana, R.
\newblock Predicting {{Good Probabilities}} with {{Supervised Learning}}.
\newblock In \emph{{{International Conference}} on {{Machine Learning}}
  (ICML)}, pp.\  625--632, New York, NY, USA, 2005.
\newblock ISBN 978-1-59593-180-1.
\newblock \doi{10.1145/1102351.1102430}.

\bibitem[Pedregosa et~al.(2011)Pedregosa, Varoquaux, Gramfort, Michel, Thirion,
  Grisel, Blondel, Prettenhofer, Weiss, Dubourg, Vanderplas, Passos,
  Cournapeau, Brucher, Perrot, and Duchesnay]{scikit-learn}
Pedregosa, F., Varoquaux, G., Gramfort, A., Michel, V., Thirion, B., Grisel,
  O., Blondel, M., Prettenhofer, P., Weiss, R., Dubourg, V., Vanderplas, J.,
  Passos, A., Cournapeau, D., Brucher, M., Perrot, M., and Duchesnay, E.
\newblock Scikit-learn: Machine learning in {P}ython.
\newblock \emph{Journal of Machine Learning Research}, 12:\penalty0 2825--2830,
  2011.

\bibitem[Platt(1999)]{Platt1999}
Platt, J.
\newblock Probabilistic outputs for support vector machines and comparisons to
  regularized likelihood methods.
\newblock \emph{Advances in large margin classifiers}, 10\penalty0
  (3):\penalty0 61--74, 1999.

\bibitem[Raposo et~al.(2017)Raposo, Santoro, Barrett, Pascanu, Lillicrap, and
  Battaglia]{Raposo2017}
Raposo, D., Santoro, A., Barrett, D., Pascanu, R., Lillicrap, T., and
  Battaglia, P.
\newblock Discovering objects and their relations from entangled scene
  representations.
\newblock In \emph{Workshops at the {{International Conference}} on {{Learning
  Representations}} ({{ICLR}})}, 2017.

\bibitem[Rubin et~al.(1984)]{rubin1984bayesianly}
Rubin, D.~B. et~al.
\newblock Bayesianly justifiable and relevant frequency calculations for the
  applied statistician.
\newblock \emph{The Annals of Statistics}, 12\penalty0 (4):\penalty0
  1151--1172, 1984.

\bibitem[{Sanchez-Gonzalez} et~al.(2018){Sanchez-Gonzalez}, Heess,
  Springenberg, Merel, Riedmiller, Hadsell, and
  Battaglia]{Sanchez-Gonzalez2018}
{Sanchez-Gonzalez}, A., Heess, N., Springenberg, J.~T., Merel, J., Riedmiller,
  M., Hadsell, R., and Battaglia, P.
\newblock Graph {{Networks}} as {{Learnable Physics Engines}} for {{Inference}}
  and {{Control}}.
\newblock In Dy, J. and Krause, A. (eds.), \emph{Proceedings of the 35th
  {{International Conference}} on {{Machine Learning}}}, volume~80, pp.\
  4470--4479. {PMLR}, July 2018.

\bibitem[Satorras \& Estrach(2018)Satorras and Estrach]{Satorras2018}
Satorras, V.~G. and Estrach, J.~B.
\newblock Few-{{Shot Learning}} with {{Graph Neural Networks}}.
\newblock In \emph{International {{Conference}} on {{Learning
  Representations}}}, 2018.

\bibitem[Sen et~al.(2008)Sen, Namata, Bilgic, Getoor, Galligher, and
  {Eliassi-Rad}]{Sen2008}
Sen, P., Namata, G., Bilgic, M., Getoor, L., Galligher, B., and {Eliassi-Rad},
  T.
\newblock Collective {{Classification}} in {{Network Data}}.
\newblock \emph{AI Magazine}, 29\penalty0 (3):\penalty0 93, 2008.
\newblock ISSN 2371-9621.
\newblock \doi{10.1609/aimag.v29i3.2157}.

\bibitem[Shchur et~al.(2018)Shchur, Mumme, Bojchevski, and
  G\"unnemann]{Shchur2018}
Shchur, O., Mumme, M., Bojchevski, A., and G\"unnemann, S.
\newblock Pitfalls of {{Graph Neural Network Evaluation}}.
\newblock \emph{arXiv:1811.05868 [cs, stat]}, November 2018.

\bibitem[Simpson(1949)]{Simpson1949}
Simpson, E.~H.
\newblock Measurement of diversity.
\newblock \emph{Nature}, 163\penalty0 (4148):\penalty0 688, 1949.

\bibitem[Sukhbaatar et~al.(2016)Sukhbaatar, Szlam, and Fergus]{Sukhbaatar2016}
Sukhbaatar, S., Szlam, A., and Fergus, R.
\newblock Learning {{Multiagent Communication}} with {{Backpropagation}}.
\newblock In Lee, D.~D., Sugiyama, M., Luxburg, U.~V., Guyon, I., and Garnett,
  R. (eds.), \emph{Advances in {{Neural Information Processing Systems}} 29},
  pp.\  2244--2252. {Curran Associates, Inc.}, 2016.

\bibitem[Vaicenavicius et~al.(2019)Vaicenavicius, Widmann, Andersson, Lindsten,
  Roll, and Sch\"on]{Vaicenavicius2019}
Vaicenavicius, J., Widmann, D., Andersson, C., Lindsten, F., Roll, J., and
  Sch\"on, T.~B.
\newblock Evaluating model calibration in classification.
\newblock In \emph{International {{Conference}} on {{Artificial Intelligence}}
  and {{Statistics}} ({{AISTATS}})}, Naha, Okinawa, Japan, 2019.

\bibitem[Veli{\v c}kovi\'c et~al.(2018)Veli{\v c}kovi\'c, Cucurull, Casanova,
  Romero, Li\`o, and Bengio]{Velickovic2018}
Veli{\v c}kovi\'c, P., Cucurull, G., Casanova, A., Romero, A., Li\`o, P., and
  Bengio, Y.
\newblock Graph {{Attention Networks}}.
\newblock In \emph{International {{Conference}} on {{Learning
  Representations}}}, Vancouver, Canada, 2018.

\bibitem[Xu et~al.(2019)Xu, Hu, Leskovec, and Jegelka]{Xu2018}
Xu, K., Hu, W., Leskovec, J., and Jegelka, S.
\newblock How {{Powerful}} are {{Graph Neural Networks}}?
\newblock In \emph{International {{Conference}} on {{Learning
  Representations}}}, 2019.

\bibitem[Yang et~al.(2017)Yang, Ribeiro, and Neville]{Yang2017}
Yang, J., Ribeiro, B., and Neville, J.
\newblock Stochastic {{Gradient Descent}} for {{Relational Logistic
  Regression}} via {{Partial Network Crawls}}.
\newblock In \emph{Proceedings of {{The}} 7th {{International International
  Workshop}} on {{Statistical Relational AI}} ({{StarAI}})}, July 2017.

\bibitem[Zadrozny \& Elkan(2001)Zadrozny and Elkan]{Zadrozny2001}
Zadrozny, B. and Elkan, C.
\newblock Obtaining {{Calibrated Probability Estimates}} from {{Decision
  Trees}} and {{Naive Bayesian Classifiers}}.
\newblock In \emph{Proceedings of the {{Eighteenth International Conference}}
  on {{Machine Learning}}}, ICML '01, pp.\  609--616, San Francisco, CA, USA,
  2001. {Morgan Kaufmann Publishers Inc.}
\newblock ISBN 978-1-55860-778-1.

\bibitem[Zadrozny \& Elkan(2002)Zadrozny and Elkan]{Zadrozny2002}
Zadrozny, B. and Elkan, C.
\newblock Transforming {{Classifier Scores}} into {{Accurate Multiclass
  Probability Estimates}}.
\newblock In \emph{Proceedings of the {{Eighth ACM SIGKDD International
  Conference}} on {{Knowledge Discovery}} and {{Data Mining}}}, pp.\  694--699,
  Edmonton, Alberta, Canada, 2002. {ACM}.
\newblock ISBN 1-58113-567-X.
\newblock \doi{10.1145/775047.775151}.

\end{thebibliography}
